# Human Body Pose Estimation for Gait Identification: A Comprehensive Survey of Datasets and Models

Human Body Pose Estimation: A Systematic Review

A Comprehensive Survey of Datasets and Models

Luke K Topham

School of Computer Science, Liverpool John Moores University, L.K.Topham@2021.ljmu.ac.uk

Wasiq Khan

School of Computer Science, Liverpool John Moores University, W.Khan@ljmu.ac.uk

Dhiya Al-Jumeily

School of Computer Science, Liverpool John Moores University, D.Aljumeily@ljmu.ac.uk

Abir Hussain

Department of Electrical Engineering, University of Sharjah, Abir.Hussain@sharjah.ac.ae

Person identification is a problem that has received substantial attention, particularly in security domains. Gait recognition is one of the most convenient approaches enabling person identification at a distance without the need of high-quality images. There are several review studies addressing person identification such as the utilization of facial images, silhouette images, and wearable sensor. Despite skeleton-based person identification gaining popularity while overcoming the challenges of traditional approaches, existing survey studies lack the comprehensive review of skeleton-based approaches to gait identification. We present a detailed review of the human pose estimation and gait analysis that make the skeleton-based approaches possible. The study covers various types of related datasets, tools, methodologies, and evaluation metrics with associated challenges, limitations, and application domains. Detailed comparisons are presented for each of these aspects with recommendations for potential research and alternatives. A common trend throughout this paper is the positive impact that deep learning techniques are beginning to have on topics such as human pose estimation and gait identification. The survey outcomes might be useful for the related research community and other stakeholders in terms of performance analysis of existing methodologies, potential research gaps, application domains, and possible contributions in the future.

# 1 INTRODUCTION

Gait refers to a person's manner of walking. Gait analysis and identification are important to several fields such as medicine and policing. By analysing a person's gait, it is possible to extract features such as velocity, cadence (steps per minute), stride duration, stride length, and stride width [153]. The combination of values for these features is unique to each individual and can therefore be considered biometric data [160], and from as early as 1977, researchers have shown its usefulness in identifying individuals [30].

Traditionally, gait-based person identification methods have used silhouette images to represent a person in an image frame, and have shown considerable success in the existing literature [49,105,129,190,196]. However, there are several limitations associated with these methods. For instance, silhouette images can lose fine-grained spatial information or they may contain other visual clues such as clothing and accessories which may influence the identification [173]. Furthermore, numerous studies do not address or explain how silhouette images are obtained in complex scenes which makes the silhouette approach unreliable for the gait recognition [173].

Alternatively, the skeleton-based approach, leverages modern advances in Human Pose Estimation (HPE) [173]. The HPE allows robust skeleton poses to be extracted directly from RGB images. Skeleton poses provide the opportunity to analyse gait more strictly by starting with a clear representation of a person in an image frame. This approach provides a more elegant extraction of gait features and the opportunity to remove the issues associated with clothing and accessories which impact the silhouette approach.

## 1.1 Existing Survey Studies

Generally, the HPE techniques can be categorised into 2D HPE and 3D HPE. In 2D HPE joints and/or body parts are tracked across the surface of an image, whereas 3D HPE also estimates the depth of the joints and body parts in the image. Table 1 summarises the existing survey studies that review 2D and 3D HPE. The survey studies presented in [31,46,118,194] focus solely on 2D HPE, [56,70,155] focus on 3D HPE, and [25,69,93,168,220] address the hybrid models. It is also important to note that previous survey studies published before 2015, review more conventional approaches to HPE, whilst [25,31,220] survey present the deep learning-based approaches to HPE. Both traditional and deep learning-based approaches are reviewed in [46,197]. Likewise, human parts parsing methods of HPE are reviewed in [93]. Furthermore, head pose estimation which is also a related challenging area of research is addressed recently [119,158]. To the best of the authors' knowledge, existing survey studies (as presented in Table 1) disregard the review of HPE for gait identification despite it enables the convenient way of gait feature extraction. Similarly, the existing surveys lack the discussion on the current reliance upon key point labelling in HPE datasets that is addressed in our study.

Table 1. A List of Existing Surveys on HPE along-with Brief Summary of Key Objectives

| Ref | Year | Description |
|---|---|---|
| [168] | 2021 | A survey of 2D and 3D multi-person HPE. Includes a focus on the unique challenges of multi-person HPE such as occlusion and truncated body parts. |
| [194] | 2021 | A survey of methods for 2D multi-person pose estimation. Includes discussion of available datasets, evaluation metrics and open-source systems. |
| [220] | 2020 | A survey on deep learning-based 2D and 3D HPE, including performance evaluation of HPE methods. |
| [25] | 2020 | A survey of deep learning-based 2D and 3D HPE between 2014 and 2020. |
| [118] | 2020 | A survey of 2D HPE models and methods between 2014 and 2020. |
| [70] | 2020 | A survey of 3D HPE from monocular RGB images. Includes a taxonomy of approaches. Includes a discussion of datasets, evaluation metrics and provides a quantitative comparison of methods. |
| [158] | 2020 | A survey of the developments of head pose estimation between 2010 and 2020. |

| Ref | Year | Description |
|---|---|---|
| [31] | 2019 | A survey on deep learning-based 2D HPE. |
| [197] | 2018 | A survey of RGB-D-based motion recognition focused on the application of deep learning to motion recognition. |
| [46] | 2016 | A survey of 2D HPE from monocular images including traditional and deep learning approaches. Includes a brief discussion of available datasets. |
| [155] | 2016 | A survey of 3D HPE from RGB images and image frames. Includes monocular and multi-view approaches. Also includes a brief discussion of available datasets. |
| [102] | 2016 | A survey of computer vision-based camera localisation including a survey of proposed extensions. |
| [93] | 2015 | A survey of 2D and 3D HPE. Includes an overview of the body parts parsing methods including monocular and multi-view approaches. Includes a brief review of available datasets. |
| [143] | 2014 | A survey of 2D and 3D human pose estimation including a general taxonomy to group model-based approaches to HPE. |
| [56] | 2012 | A survey of multi-view approaches for 3D HPE and activity recognition. |
| [69] | 2010 | A survey of view-invariant pose representation and estimation using a marker-less approach. Includes an overview of behaviour analysis. |
| [119] | 2009 | A survey of head pose estimation methods including a taxonomy of head pose estimation methods. |
| [155] | 2007 | A survey of human motion analysis using a marker-less approach. |
| [113] | 2006 | A survey of computer vision-based motion capture tracking, pose estimation and recognition between 2000 and 2006. |
| [112] | 2001 | A survey of computer vision-based motion capture including tracking, HPE and recognition. |

Table 2. A List of Existing Surveys on Gait Analysis/Identification along-with Brief Summary of Key Objectives

| Ref | Year | Description |
|---|---|---|
| [157] | 2022 | A survey of deep learning approaches to gait recognition. Includes a four-dimensional taxonomy and a review of datasets. |
| [41] | 2022 | A survey of deep learning approaches to gait recognition. Includes a review of datasets, approaches, and architectures. Highlights weaknesses of the approach. |
| [80] | 2021 | A survey of vision-based approaches to gait recognition, focusing mainly on silhouette images. |
| [74] | 2021 | A survey of vision-based approaches to gait recognition. Includes a review of gait features, datasets, and existing solutions. |
| [132] | 2019 | A survey of publicly available depth-based gait datasets for person identification and/or classification. |
| [103] | 2019 | A survey of the use of gait for person identification via the use of wearable sensors. |
| [35] | 2019 | A review of the evidence demonstrates a relationship between gait, emotions, and mood disorders. |
| [48] | 2018 | A survey of gait analysis for human identification. |
| [192] | 2018 | A survey of gait recognition methods, including a range of sensors. Includes a discussion of machine learning approaches. |
| [166] | 2018 | A survey of vision-based gait recognition methods. Includes a discussion of evaluation metrics. |
| [29] | 2018 | A survey of the types of features used by different modalities of gait recognition. Includes a discussion of factors that impact gait recognition and the issues of gait spoofing and obfuscation. |
| [43] | 2007 | A survey of biometric recognition via the use of gait analysis. Includes a discussion of factors that influence gait recognition. In addition, an evaluation of gait analysis under various attack scenarios are presented. |

Table 2 lists the existing survey studies on the topics of computational gait analysis and gait-based person identification. The survey study in [132] reviews the depth-based gait datasets. In addition to person identification, there is a range of other domains in which gait analysis has the potential applications, for example, the association between gait and emotion is explored in [35]. In addition to applications, gait-based identification approaches can also be categorised w.r.t sensors used in the approach. The main sensors and approaches used for gait identification are machine vision, floor sensors, and wearable sensors. Among the list in Table 2, [120,166] solely reviews computer vision techniques, [103] reviews only wearable sensors methods, and [29,35,43,192] review approaches with a variety of sensors. A review of machine learning and deep learning approaches for gait identification is presented in [192] and [41,157] respectively. A discussion of the factors that limit the accuracy of gait identification and the security challenges are provided in [29,43].

Alternatively, silhouette-based approaches to gait identification are surveyed in [35,48,120,166], while [103,192] present the approaches using sensors other than vision-based (e.g., wearable sensors). To the best of the authors' knowledge, none of the existing surveys present the review of the skeleton-based approach to gait identification. Furthermore, the existing surveys lack the review of the HPE datasets, models, and methods available for skeleton-based gait identification.

### 1.2 Motivation

The development of the skeleton-based gait identification approaches has the potential impact within the diverse application domains such as physiotherapy, sports coaching, healthcare, and person or criminal identification [55,133]. For example, the police and security services could benefit from person identification systems that are not dependent on appearance, such as face identification, which can otherwise be easily disguised.

Recently, BBC reported the failure of existing face matching tools that are deployed by UK police are inaccurate [42]. Big Brother Watch, a campaign group investigated the technology and found that it produces an extremely high number of false positives, therefore inappropriately identifying innocent people as suspects. Likewise, 'INDEPENDENT' [159] and 'WIRED' [18] reported that Metropolitan and South Wales Police's facial recognition technology misidentify the suspects, with a false positive rate of 98%. These statistics indicate the existence of a major gap within the existing technology that is needed to be investigated to deal with challenges associated with real-time video processing.

Improving the real-time person identification has the potential impact in various domains such as border security at airports, police enforcement, and domestic applications (e.g., prevention of thefts in shops). The recent improvements in gait-based identification have shown to be applicable in situation where face-matching technology shows poor performance, for example, with low-resolution images, and/or when the face is covered. Furthermore, skeleton-based gait identification has indicated success in producing distinct features for the identification of individuals, such as stride length, stride width, and gait energy. Hybrid methods combining the use of face matching technology with gait-based identification may provide higher identification accuracy.

In addition to the aforementioned impacts and related methods, the existing survey studies lack the comprehensive review of skeleton-based approaches to gait identification, and the available datasets for this problem. Similarly, application domains are not thoroughly reviewed. This survey therefore should support the understanding of gait-based identification methods to aid in the development of such solutions, including the collection of more appropriate datasets in future, with the aim of improvements towards safer community.

### 1.3 Contributions

For the first time, we present a thorough review of HPE methods specifically aimed towards gait identification, with an emphasis on skeleton-based approaches. This survey aims to address the shortcoming of the previous surveys by providing a systematic review of HPE for skeleton-based gait identification. Furthermore, we present a review of 2D and 3D HPE methods and gait identification methods, including a discussion of datasets, limitations in the existing datasets and technical models, application domains, and evaluation metrics. This survey mainly contributes in the following aspects:

- A comprehensive review of approaches to the problems of 2D HPE, gait analysis, and gait identification.
- A comprehensive review of the skeleton-based approaches to gait analysis with a comparison to conventional methods.
- A comprehensive review of the datasets available for the training and evaluation of HPE solutions and gait-based identification, including a summary of their limitations and suggestions for future datasets.
- An overview of the potential applications of HPE and gait analysis followed by suggested future works.

This paper describes and compare the traditional silhouette-based and the modern skeleton-based approaches to gait analysis and person identification. We also present a comprehensive review of the existing HPE technologies and techniques required to implement the skeleton-based approach, along with a review of the datasets and evaluation metrics that have been used to implement and evaluate HPE. Finally, this paper highlights a diverse range of domains in which gait analysis and HPE have the potential to make a significant impact, with a particular focus on the person identification.

### 1.4 Organisation

The remainder of this paper is organised as follows. Section 2 addresses the methodology of the proposed survey study. Section 3 reviews a wide range of datasets for HPE and gait identification in subsections 3.1 and 3.2 respectively. Section 4 presents HPE, the metrics used to evaluate and compare methods, and technical approaches. Section 5 introduces the problem of gait identification. The traditional silhouette-based approaches are reviewed in subsection 5.1, while the more modern skeleton-based approach based on HPE is reviewed in subsection 5.2. Domains and applications relevant to HPE and gait analysis are explored in section 6. A discussion of the major topics is provided in section 7. Finally, section 8 provides the conclusions and suggested improvements in various aspects for future works based on our findings from this study.

## 2 METHODOLOGY

This survey aims to systematically review existing research in 2D and 3D HPE to aid future work in HPE and gait identification in various aspects as mentioned earlier. The following subsections explains the adapted methodology for this review. The scope of this article is guided through two filters: research aspects and search strategy.

### 2.1 Research Aspects

The research aspects investigated in this survey include: **a)** What datasets are available for HPE and Gait identification? What are their limitations? **b)** What models are available for HPE and Gait identification? What are their limitations? **c)** What possible improvements can be made to the datasets and models? **d)** Is there a significant difference in the results of the skeleton-based approaches to gait identification as compared to traditional approaches?

The survey firstly focuses on the identification and evaluation of the datasets required to evaluate the different methods of HPE and gait identification. The datasets also provide the opportunity for training and evaluating machine learning-based approaches to HPE and gait identification. Secondly, we aim to represent the variety of approaches to HPE and gait identification, to evaluate them, and to explore the relationships between them. We then identify various aspects of datasets and models which can be explored further or improved in future works. The study also explores the works relevant to the skeleton-based approaches to gait identification where we perform statistical tests to investigate the performance measures.

### 2.2 Search Strategy

Inspired by [168], Table 3 includes a list of keywords and permutations used to explore the HPE literature for this survey. The keywords are categorised into context and objective. Context refers to the HPE methods as required to achieve the objectives of the HPE elements of this survey. Objective refers to the objective of identifying body parts. Table 3 contains a list of permutations used within the set of keywords when inputting search phrases. The following database libraries were searched for this survey: IEEE Xplore, Science Direct, ACM Digital Library, Scopus, and Google Scholar (see Figure 1). To filter the vast research and find only relevant studies aligned with the objectives of this survey, we defined a range of selection and quality assessment criteria that considers: a) only articles written in English; b) only articles published in peer-reviewed journals or conferences are considered to ensure the quality of research; c) non-repeated articles.

Table 3. Keywords Used to Explore HPE Literature for this Survey

| Goal | Keywords |
|---|---|
| Context | Human, Body, Human Body, Part-based, Model-based, Model-Free, Regression-based, Top-Down, Bottom-Up |
| Objective | Detection, Identification, Estimation, Recognition, Recovery |
| Permutations | [(Context AND "Pose") OR (Objective + Context + "Pose")] AND ("Estimation" OR Tracking) |

Using the above search strategy, a variety of peer-reviewed studies are identified. Figure 1 shows that 31.9% of the papers reviewed are journal papers while the remaining 68.1% are conference papers. Figure 1 also shows the distribution of publishers where, publishers who account for 3% or less are categorised in 'other', this includes MIT, ACM (journal), AAAI, and CVPR.

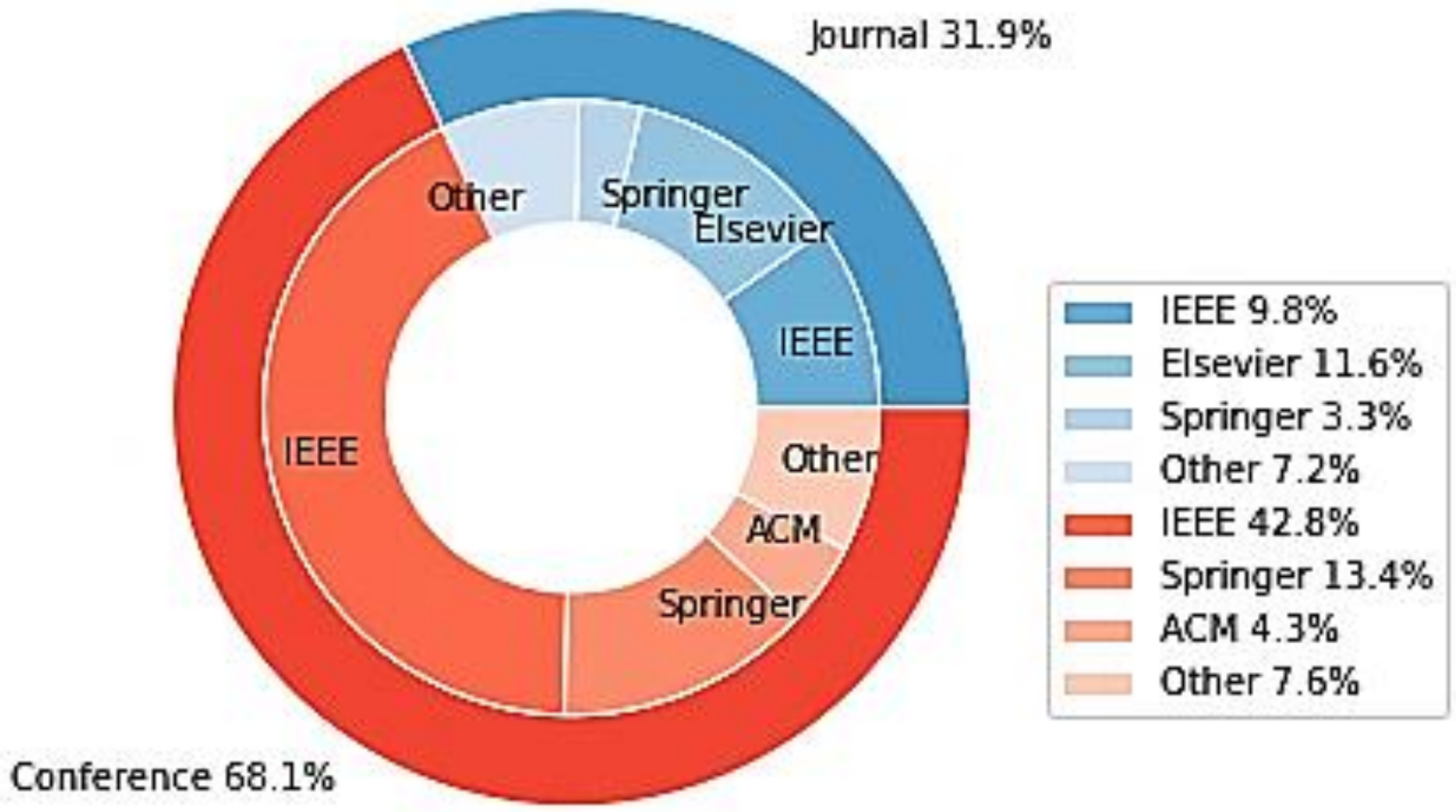

Figure 1. Distribution of Publication Types Used in this Study

Using the above search strategy, a variety of peer-reviewed journal and conference papers were found. FIGURE 1 shows that 31.9% of the papers reviewed were journal papers while the remaining 68.1% are conference papers. Figure 1 also shows the proportion of publishers, publishers which account for 3% or less are categorised in 'other', that includes MIT, ACM (journal), AAAI, and CVPR.

## 3 DATASETS

To develop and evaluate the HPE and gait recognition approaches, a dataset (labelled, unlabelled) is the primary requirement. Collecting, compiling, and labelling datasets is a time-consuming process and therefore, many studies leverage secondary datasets published in the literature. This section will provide a systematic review of existing datasets available for training and evaluating models for 2D HPE, 3D HPE, and gait identification. For HPE, single image frames are often sufficient, however, for gait identification video data of a complete gait cycle is required. Some datasets are larger in size and contain features for multiple tasks; therefore, they may appear in more than one table. Though most datasets include RGB, RGB-D, or motion capture data for HPE, some datasets for gait identification include data from devices such as wearable sensors, microphones and pressure sensors.

### 3.1 Datasets for Human Pose Estimation

This section will review the datasets available for both 2D and 3D HPE, as listed in Table 4. Common trends and limitations will also be highlighted in the extended in the supplementary materials Table S5. Table 4 provides a summary of the current datasets available for both 2D and 3D HPE. It can be noticed the number of records contained in each dataset varies, ranging from 928 in [68] to 5.4 million records in [199]. In relation to labelling methods, majority of the 2D datasets include labelled keypoints comprising 2D coordinates of body joints in an image that are grouped for each person. As most of the 3D datasets are captured in motion capture laboratories, these usually provide 3D coordinates of the reflective markers worn by each person in the images. In addition to key point labelling, some datasets also provide semantic segmentation of objects and body parts.

Table 4. Summary of Existing Datasets Available for 2D and 3D HPE

| Ref | Dataset Name | No. Records | Labelling | 2D/ 3D |
|---|---|---|---|---|
| [90] | Common Objects in Context (COCO) | 200,000 | Key points. | 2D |
| [3] | MPII Human Pose | 40,000 | Key points, full 3D torso, and head orientation, occlusion labels for joints and body parts, and activity labels. | 2D |
| [62] | Human3.6M | 3,600,000 | Motion capture. | 2D & 3D |
| [205] | PASCAL-Person-Part | 3,533 | Key points and body segmentation. | 2D |
| [67] | Buffy2 Stickmen, Movie Stickmen | ~6,000 | Key points. | 2D |
| [154] | Frames Labeled in Cinema (FLIC) | 20,928 | Key points. | 2D |
| [189] | Synthetic hUmans foR REAL tasks (SURREAL) | 67,582 | Ground-truth pose, depth maps, and segmentation masks. | 3D |
| [165] | HumanEva | 40,000 | 3D Markers, 2D position of Markers. | 2D & 3D |
| [108] | MuCo-3DHP | 8,000 | Marker-less motion capture. Occlusion annotation. | 3D |
| [72] | PANOPTIC Studio | 297,000 | 3D Pose. | 3D |
| [101] | 3D Poses in the Wild | 51,000 | 2D Pose and 3D Pose. | 2D & 3D |
| [71] | Leeds Sports Pose Extended | 10,000 | Key points. | 2D |
| [37] | Joint Track Auto (JTA) | 500,000 | Key points, 3D Pose. | 2D & 3D |
| [106] | MPI-INF-3DHP | >1,300,000 | 3D Pose. | 3D |
| [199] | CSI | 5,400,000 | 3D Pose. | 3D |
| [218] | Actemes | 2,326 | Key points. | 2D |
| [68] | Joint labelled Human Motion DataBase (J-HMDB) | 928 | Key points. | 2D |

Despite the availability of variety datasets (as in Table 4), the quantity and diversity (e.g., unusual poses) is limited within these datasets. A common issue with HPE (which is discussed later in this study) is its difficulty in dealing with unusual poses. Unusual poses may include a person being upside down, or dynamic perspective (e.g., yoga exercise). However, the LSPE dataset includes images from a variety of sports where some of the poses in this dataset can be considered unusual [71].

**Error! Reference source not found.** presents an overview of several properties associated with the datasets listed in Table 4, including the number of viewpoints captured, the experimental environments, data labelling method, and occlusion labelling. The distribution helps in the identification of some limitations in the current datasets. For example, **Error! Reference source not found.** shows that only 9% of the HPE datasets reviewed contain multiple viewpoints, completely missing the 2D HPE. Approximately 42% of the datasets do not contain a variety of indoor and outdoor locations and only 4% of all datasets are 3D HPE datasets containing outdoor images (see **Error! Reference source not found.**). The distribution of labelling types indicates that 84% of all HPE datasets contain only key point or marker-based labelling. The presence of occlusion labelling indicates that only 41% of datasets reviewed contains occlusion labelling. This is particularly important as occlusion is a significant problem in HPE and resolving this problem may provide more reliable gait data for gait identification. Furthermore, the majority of the 2D datasets presented in **Error! Reference source not found.** are labelled with key points, usually the body joints. Microsoft's COCO dataset is one of the most popular datasets for 2D Pose estimation [91]. This is a large-scale object detection, segmentation, and captioning dataset and includes 91 different object categories in 330,000 images [90]. The dataset includes challenging and uncontrolled conditions in both indoor and outdoor environments. The COCO dataset has been used to train several high-profile 2D HPE methods such as OpenPose [20] and PersonLab [140].

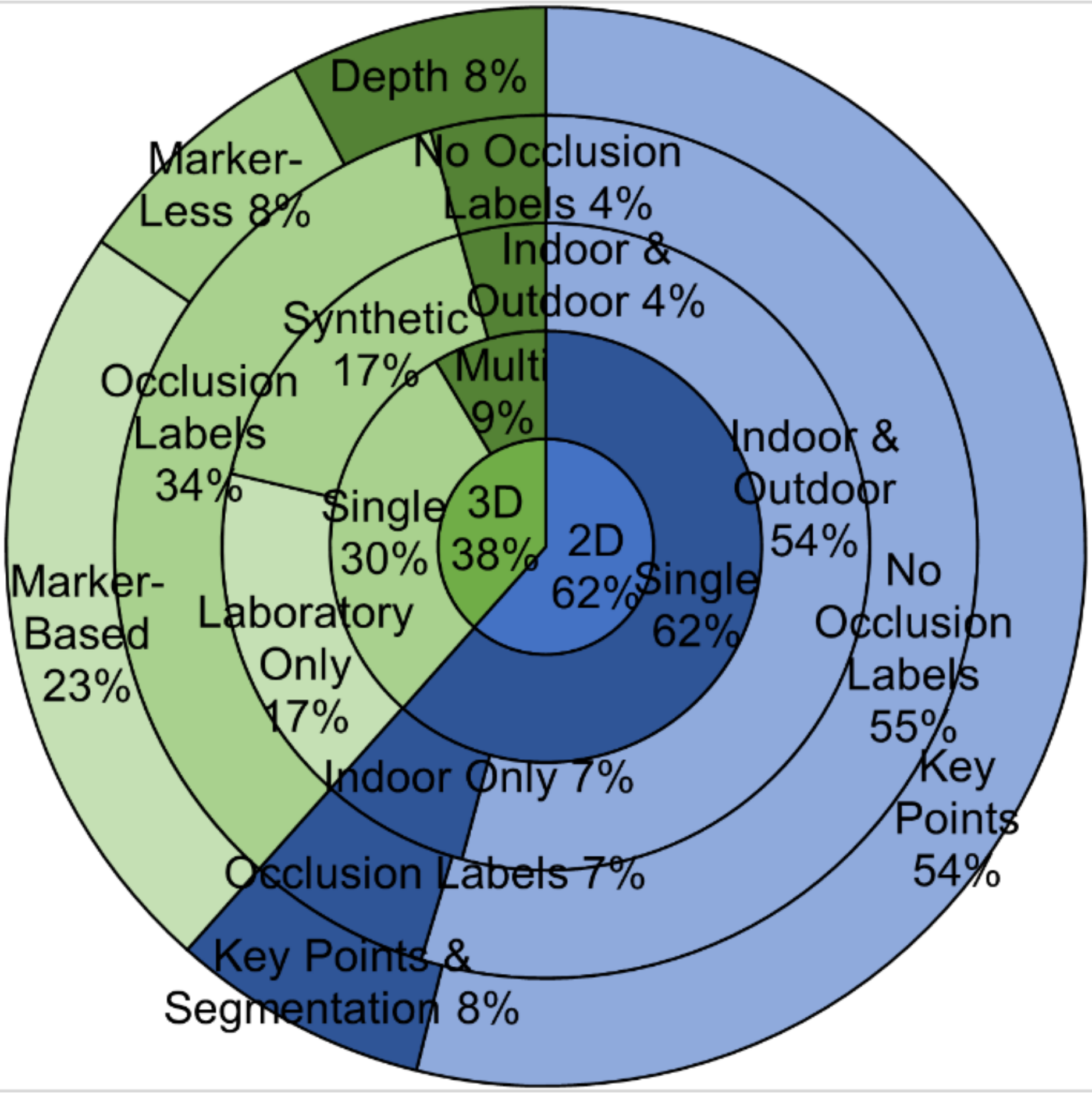

Figure 2. HPE Dataset Properties in the Works Reviewed including Viewpoints, Environment, Labelling, and Occlusion Labelling

In comparison, most of the 3D datasets are recorded in motion capture suites as listed in Table 4. While the 3D datasets are ordinarily restricted to an indoor environment such as a laboratory, green screens and synthetic images have been used in some cases to increase the diversity in background, such as in [62,106]. MuPots-3D is an exception to this as the authors used marker-less motion capture in outdoor environments [108]. HPE algorithms enable the construction of skeleton models from the key points and marker data [143,197]. While motion data such as optical flow can provide useful information for HPE [26], many of the datasets, such as COCO [91], Pascal-Person-Part [205], VGG [67], and FLIC [154], primarily include single-frame images, therefore, it is not possible to extract motion information. To partially overcome this, the MPII dataset includes adjacent video frames to allow the extraction of motion information [3].

Due to the nature of human pose datasets, it can be difficult to guarantee the privacy to participants in the images. Therefore, some datasets contain only pre-processed images that provide only derived or silhouettes images such as in [10,23,190,196], as shown in Table 5. Synthetic images provide an opportunity to create data using partially or fully computer-generated images [127]. The manual labelling of data is a time-consuming task and therefore, the creation of synthetic images can help to automate this process. The Human3.6M dataset includes some mixed-reality images to increase the number of records in the dataset [62], and another dataset for 3D HPE presented in [189], SURREAL, is completely composed of synthetic images. Similarly, the JTA dataset is primarily composed of images taken from video games [37]. However, one limiting factor of the adoption of such datasets is validation of associated models (that are being trained over such synthetic datasets) in real-time settings. Overcoming this challenge may aid in the adoption of synthetic data, therefore increasing the amount of data available.

Furthermore, existing datasets suffer from additional limitations, for instance, most of the datasets do not reflect real-time environmental variations, with being confined to laboratories or simulated settings [165,199]. On the other hand, some datasets are labelled either incompletely or inconsistently [44,57]. Likewise, several datasets contain poor-quality, low-resolution images [221]. Although HPE and gait recognition are affected by the viewing angle, viewing angles are rarely labelled in the few datasets that provide footage from multiple angles [81,212].

Another important factor in HPE and gait recognition datasets is clothing (wearing) which also affects the HPE and gait recognition accuracy [88]. However, very few datasets are available that include subjects in a variety of clothing, such as those presented in [106,108]. In addition, datasets recorded using marker-based motion capture restricts participants to skin-tight clothing with markers attached [106]. One alternative would be to use a marker-less system as presented in [72], allowing participants to wear a variety of clothing. However, this requires an expensive setup involving a purpose-built dome containing hundreds of cameras. Furthermore, the datasets listed in Table 4 do not contain a significant number of heavily crowded images. This ultimately leads to poor HPE model training and therefore affects the performance specifically, in real-time highly overlapped circumstances

### 3.2 Datasets for Gait Identification

This section will highlight the datasets available for the gait identification. Table 5 highlights some of the most popular datasets in this area, including data from a range of sensors such as cameras, motion capture, and inertial sensors. An extended table in the supplementary materials, Table S6, highlights some of the advantages and disadvantages of the datasets. For a dataset to be useful for the problem of gait identification, it must contain multiple gait cycles per subject, preferably from separate walking exercises, provided by multiple subjects. Capturing multiple real-world environments, a variety of clothing, and footwear for the participants might enable the generalisation of the machine learning models being trained. Some of the datasets outlined in Table 5 were created for other domains such as healthcare however, may be useful for the gait identification. Though, it would first be necessary to ensure that the datasets include sufficient complete gait cycles per participant. It may be difficult to find a sufficient number of complete gait cycles where there are a limited number of records per participant, such as in [6,7,53]

Table 5. Comparison of Existing Datasets that are Available for the Gait Identification

| Ref | Dataset Name | No. Subjects | No. Records | Device Type |
|---|---|---|---|---|
| [190] | IST gait database | 21 | 72 | Digital Camera |
| [33] | Human Gait Phase Dataset | 21 | 35,306 | Motion Capture. |
| [186,187] | Human Gait (walking) Database | 93 | 334 | Phone Accelerometer. |

| Ref | Dataset Name | No. Subjects | No. Records | Device Type |
|---|---|---|---|---|
| [27,28] | HuGaDB: Human Gait Database | 18 | 2,111,962 | Inertial Sensors. |
| [6,7] | Gait Sounds | 55 | 55 | Audio. |
| [23] | Gait Silhouette Dataset | 294 | 2,940 | RGB Camera. |
| [131,133] | Gait Recognition Image and Depth Dataset (GRIDDS) | 35 | 350 | RGB-D Camera. |
| [55] | TUM Gait from Audio, Image and Depth (TUM-GAID) | 305 | 3,370 | RGB-D, Microphone. |
| [163,183] | SOTON HiD | 100 | 2,280 | RGB Camera. |
| [160] | EMOGAIT | 60 | 1,140 | RGB Camera. |
| [196] | Unnamed | 20 | 240 | RGB Camera. |
| [212] | CASIA Gait Database: Dataset B | 124 | 13,640 | RGB Camera. |
| [156] | A multimodal dataset of human gait at different walking speeds established on injury-free adult participants | 50 | 1,143 | Motion capture, force plates, wireless EMG. |
| [116] | An elaborate data set on human gait and the effect of mechanical perturbations | 15 | 70,000 | Motion capture, force plates, accelerometer. |
| [172] | OU-ISIR gait database, multi-view large population dataset (OU-MVLP) | 10,307 | 144,298 | RGB Camera. |
| [138] | KinectREID | 71 | 483 | RGB-D Camera. |
| [221] | Motion Analysis and Re-identification Set (MARS) | 1,261 | ~20,000 | RGB Camera. |
| [176,200] | PKU HumanID | 18 | 216 | RGB Camera. |
| [53] | Multi-shot Dataset | 200 | 400 | RGB Camera. |
| [32] | Reidentification Across indoor-outdoor Dataset (RAiD) | 43 | 6,920 | RGB Camera. |
| [104] | Wide area scenario dataset | 70 | 4,786 | RGB Camera. |

Figure S1 of the supplementary materials presents the distribution of the sensor types used to collect the primary data in Table 5. Only the image-based and motion capture datasets are appropriate for the skeleton-based approach to gait identification, which is the focus of this paper. This means that 70.8% of the datasets reviewed in this survey are appropriate for the skeleton-based approach to gait identification. One of the major limitations with image datasets is privacy and ethical concerns. Therefore some datasets, such as [23], opt to only provide the processed images and not the original RGB images, which prevents the use of the skeleton-based approach. Similarly, not all motion capture datasets include accompanying RGB images.

In addition to privacy and ethical concerns, there are several common limitations related to gait recognition datasets. Firstly, availability and access to datasets is one the common challenges for researchers. For instance, several datasets are difficult to gain access to and require a formal application which is often time-consuming. Likewise, some datasets are described but are not made public, or are removed from public access [55,160]. Secondly, limited diversity in datasets is a major concern to be highlighted. Often participants within the datasets include work colleagues, students, and social contacts, etc., producing limited diversity in respect to ethnicity, age, gender, height, and weight [28,33]. Lack of diversity in datasets affects the machine-based models' training and generalisation. Thirdly, the dynamic real-time environment and background are significant factors of image datasets. There are very few datasets considering variations of real-world outdoor and indoor environments [33,55,133]. However, generalisation of machine-based models might require comparatively larger-sized datasets comprising varying real-time dynamics. Finally, many datasets are recorded from a single angle (perspective) [133,183], and others are inconsistent with their viewing angle recording [212]. Where there are multiple viewing angles, most datasets do not label viewing angles, as can be found in [138].

Despite the existing datasets have shown to be adequate for developing gait identification algorithms and for the training of machine-based intelligent models, the aforementioned limitations should be considered while capturing datasets. Some major recommendations that may be considered while capturing datasets might include: **i)** Larger representation of participants considering data diversity (e.g., ethnicity, age, gender, height, weight), **ii)** Include a range of clothing and accessories for each participant, **iii)** Multiple complete gait cycles per participant, **iv)** Record in multiple real-world locations (indoors and outdoors), **v)** Contain RGB data, **vi)** Record from multiple angles, annotate and publish the dataset.

## 4 HUMAN BODY POSE ESTIMATION

The HPE is a fundamental computer vision problem aiming to extract the posture of human bodies from input images by estimating the locations and connections of body segments [25]. This section will review the methods and approaches for both 2D and 3D HPE. This section outlines the traditional taxonomy of 2D and 3D HPE as displayed in Figure 3 [25,31,220]. The section also presents an overview of current 2D and 3D HPE methods, including the datasets used for training and evaluation of methods as well as corresponding metrics used for performance evaluation.

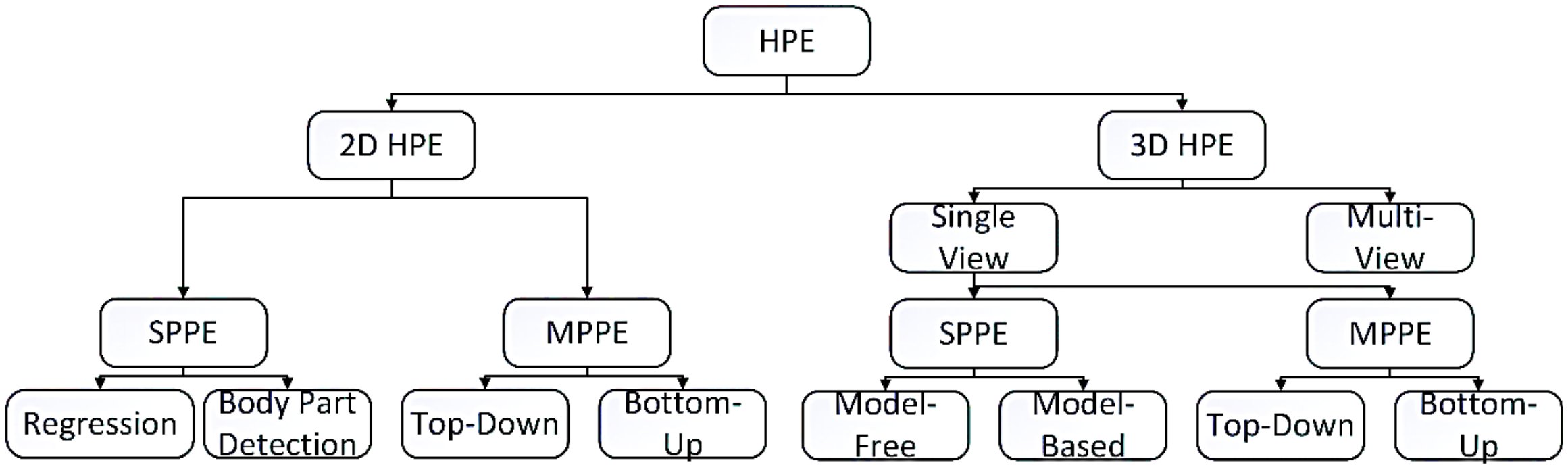

Figure 3. Taxonomy of HPE Approaches

### 4.1 Evaluation Metrics

Various evaluation metrics have been utilised to evaluate HPE models and perform comparative analysis including; Intersection over Union (IoU), Mean Pixel IoU (mIoU), Average Precision (AP), Mean Average Precision (mAP), Percentage Correct Parts (PCP), Percentage of Detected Joins (PDJ), Percentage of Correct Key points (PCK), Head-Normalized Probability of Correct Key point (PCKh), 3D Percentage of Correct Key points (3DPCK), Area Under the Curve (AUC), Object Key point Similarity (OKS), Average Distance of Key points (ADK), Mean Per Join Position Error (MPJPE), Normalised Mean Per Join Position Error (N-MPJPE), Proscrutes analysis Mean Per Join Position Error (P-MPJPE), and Mean Per Joint Angular Error (MPJAE). A complete description of each of the evaluation metrics can be found in Section 1.2 of the supplementary materials.

### 4.2 2D Human Pose Estimation

The 2D HPE aims to extract estimated human body postures from an input image or video frame. Generally, the 2D HPE can be categorised into Single Person Pose Estimation (SPPE) and Multi-Person Pose Estimation (MPPE). As the name indicates, SPPE extracts a pose from only a single person per image, where MPPE can extract poses of multiple subjects in an image. The top-down approach to MPPE often implements SPPE where first object detection is used in the first step to extract individuals from an image, and then SPPE is used on each of the extracted individuals. The left-side tree in Figure 3 shows taxonomy of 2D HPE methods, which are explained in further detail in the following subsections. As can be seen in Figure 3, 2D HPE has two top-level categories including SPPE and MPPE. The 2D SPPE can be further divided into regression-based and body part detection-based approaches while the 2D MPPE category can be divided into the top-down and bottom-up approaches. The top-down approach to MPPE often implements SPPE where first object detection is used in the first step to extract individuals from an image, and then SPPE is used on each of the extracted individuals. Table 6 summarises the existing studies addressing 2D HPE methods. Each method includes the relevant type and approach as per the taxonomy found in Figure 3.

Table 6. A Comparison of 2D HPE Methods and Models

| Ref | Year | Dataset | SPPE / MPPE | Approach | Model | Metric | Performance Score |
|---|---|---|---|---|---|---|---|
| [20] | 2021 | COCO, MPII | MPPE | Bottom-Up | CNN | mAP | 75.6% |
| [52] | 2020 | COCO | MPPE | Bottom-Up | FCN | mAP | 71.4% |
| [19] | 2020 | COCO, MPII | MPPE | Top-Down | CNN | mAP | 79.2% |
| [5] | 2020 | LSP, MPII | SPPE | Body Part Detection | CNN | PCP<br>PCK@0.2<br>PCKh@0.5 | 72.8%<br>94.5%<br>92.7% |
| [58] | 2020 | COCO | MPPE | Top-Down | Encoder - Decoder | mAP | 76.5% |
| [214] | 2020 | MPII, COCO | MPPE | Top-Down | Encoder - Decoder | mAP | 76.2 |
| [137] | 2019 | COCO | MPPE | Bottom-Up | CNN | mAP | 42.8% |
| [169] | 2019 | COCO, MPII | MPPE | Bottom-Up | CNN | mAP<br>PCKh@0.5 | 77%<br>92.3% |
| [215] | 2019 | MPII, LSP | SPPE | Regression | CNN | PCKh@0.5<br>AUC | 91.1%<br>65.9% |
| [98] | 2019 | LSP, MPII | SPPE | Regression | CNN | PCKh@0.5 | 89.1% |
| [34] | 2019 | FLIC | SPPE | Body Part Detection | CNN | PCK@0.2 | 96.4% |
| [114] | 2019 | COCO | MPPE | Top-Down | DNN | OKS AP | 73.6% |
| [78] | 2019 | COCO | MPPE | Bottom-Up | CNN | mAP | 66.7% |
| [126] | 2019 | MPII, COCO, PASCAL | MPPE | Bottom-Up | CNN | mAP | 78.5% |
| [2] | 2018 | MPII, COCO | MPPE | Top-Down | CNN | mAP | 68.7% |
| [128] | 2018 | LSP, MPII | SPPE | Regression | DNN | PCKh@0.5 | 91.2% |
| [140] | 2018 | COCO | MPPE | Bottom-Up | CNN | mAP | 68.7% |
| [97] | 2018 | Penn Action, JHMDB, LSP, MPII | SPPE | Body Part Detection | CNN<br>RNN | PCK@0.2 | 93.6% |
| [206] | 2018 | COCO | SPPE | Body Part Detection | CNN | mAP | 76.7% |
| [40] | 2018 | MPII | MPPE | Bottom-Up | CNN | PCKh@0.5<br>mAP | 85.6%<br>75.1% |
| [59] | 2017 | CUB-200-2011, LSP, COCO | MPPE | Top-Down | CNN | PCK | 84.5% |
| [39] | 2017 | MPII, COCO | MPPE | Top-Down | CNN | mAP | 76.7% |
| [205] | 2017 | PASCAL | MPPE | Top-Down | FCRF<br>FCN | mAP<br>mIoU<br>ADK | 39.2%<br>64.39%<br>40.7% |
| [13] | 2017 | MPII, LSP | SPPE | Body Part Detection | CNN<br>RNN | PCKh@0.5<br>AUC | 88.1%<br>58.8% |
| [209] | 2017 | LSP, MPII | SPPE | Body Part Detection | DCNN | PCKh@0.5 | 92% |
| [60] | 2017 | MPII | MPPE | Bottom-Up | CNN | mAP | 73.3% |
| [123] | 2017 | MPII, COCO, PASCAL | MPPE | Bottom-Up | Stacked Hourglass CNN | mAP | 77.5% |
| [21] | 2016 | MPII, LSP | SPPE | Regression | CNN | PCP | 81% |
| [124] | 2016 | FLIC, MPII | SPPE | Body Part Detection | CNN | PCKh@0.5 | 90.9% |
| [144] | 2016 | LSP<br>MPII | MPPE | Bottom-Up | CNN | PCP<br>PCK | 86.5%<br>63.6% |

| Ref | Year | Dataset | SPPE / MPPE | Approach | Model | Metric | Performance Score |
|---|---|---|---|---|---|---|---|
| [61] | 2016 | LSP, MPII | MPPE | Bottom-Up | CNN | mAP<br>PCK<br>mAP | 54.1%<br>66.1%<br>70% |
| [201] | 2016 | MPII, LSP, FLIC | SPPE | Body Part Detection | CNN | PCK | 90.5% |
| [89] | 2016 | MPII, LSP | SPPE | Body Part Detection | CNN | PCKh@0.5 | 89.4% |
| [17] | 2016 | MPII, LSP | SPPE | Body Part Detection | CNN | PCP<br>PCKh@0.5<br>PCK | 84.3%<br>89.7%<br>90.7% |
| [45] | 2016 | MPII, Penn Action | SPPE | Body Part Detection | CNN<br>RNN | PCKh@0.5<br>PCK | 86.1%<br>91.8% |
| [65] | 2015 | FLIC | MPPE | Bottom-Up | CNN | PDJ | 87% |
| [38] | 2015 | LSP, LSPE, FLIC | SPPE | Regression | CNN | PCP | 84% |
| [178] | 2015 | FLIC, MPII | SPPE | Body Part Detection | CNN | PCKh@0.5 | 82% |
| [180] | 2014 | FLIC, LSP | SPPE | Regression | DNN | PCP | 61% |
| [85] | 2014 | FLIC, Buffy Stickmen | SPPE | Regression | CNN | PCP | 75.27% |
| [146] | 2014 | LSP, FLIC | SPPE | Body Part Detection | DNN | PCP | 72% |
| [154] | 2013 | FLIC, Buffy PASCAL | SPPE | Body Part Detection | SVM | PDJ | 70% |

**Error! Reference source not found.** shows the proportions of each method present in the literature reviewed in this survey as summarised in Table 6 for both 2D HPE (a) and 3D HPE (b). The bar chart presented in Figure S2 of the supplementary materials displays the accuracies of the works presented in Table 6 in mAP and Figure S3 shows PCKh@0.5. In both cases, the higher the bar the better the accuracy. The best performing works were [5,20] which are discussed in the following subsections.

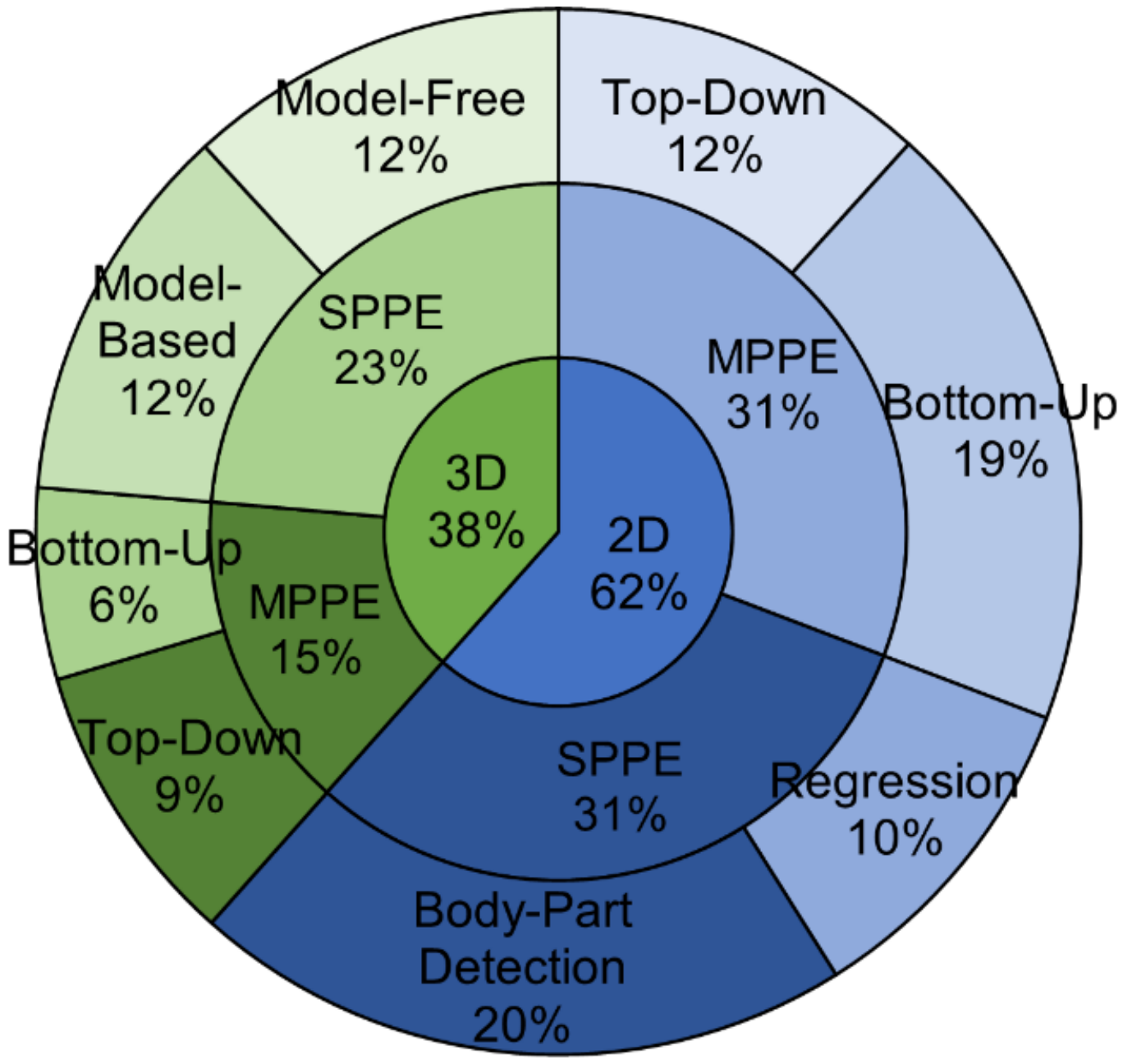

Figure 4. The Proportions of HPE Types Reviewed in this Study (a) 2D HPE, (b) 3D HPE

#### 4.2.1. *The 2D Single Person Pose Estimation*

As mentioned previously, the 2D SPPE aims to estimate the pose of a single person per input image. In cases where multiple persons are present in an image, the image is automatically cropped so that only one person is present in each sub-image, then SPPE can proceed on each sub-image. To extract the single-person sub-images, a full-body detector such as [52,148], or an upper-body detector such as [110] have been utilised. Generally, 2D SPPE are divided into two categories; regression, and body part detection, as described in the following subsections.

##### 4.2.1.1 *Regression*

Regression methods employ an end-to-end framework to automatically learn the mapping from an input image to body joints or the parameters of the body model [180]. Table 6 presents several regression-based approaches to 2D SPPE such as [21,98,128,215]. The progress of regression approaches, as with other approaches, is being accelerated by deep learning methods. Table 6 shows that in recent years, deep learning approaches, particularly Convolutional Neural Networks (CNNs) have become increasingly popular for the 2D SPPE task. The switch to deep learning approaches has accelerated the improvement in accuracy up to 79.2% mAP. DeepPose [180] is a cascaded deep neural network regressor with the ability to identify key points within the input images. For image classification, DeepPose bases its architecture on the DNN on the work by Krizhevsky et al. [79]. Carreira et al. propose an Iterative Error Feedback (IEF) network [21] which uses the GoogLeNet [170]. The IEF is a self-correcting network that feeds the estimation error back into the input [220]. Luvizon et al. [98] propose an end-to-end regression approach using soft-argmax to convert feature maps directly into body joint coordinates, resulting in a fully differentiable framework. This method can learn heat map representations indirectly, without the need to generate artificial ground truth.

Table 6 further indicates that the best performing 2D SPPE in the regression category is FPD [215] with a PCKh@0.5 of 91.1%. Although satisfactory levels of performance have been achieved using the regression-based approach, it does not outperform the body part detection methods, as discussed in the following subsection. A major factor that limits the performance of the regression-based approach is that current solutions do not exploit the structural information contained in the human body pose [87].

##### 4.2.1.2 *Body Part Detection*

The body part detection method aims to train a body part detector to estimate the locations of body joints [220]. This typically consists of two stages, first heatmaps are generated for key points, followed by the assembling of the estimated key points into body poses. Much of the recent work has considered body part detection as a heatmap detection problem [2,19,20,137]. To assemble the detected body parts into a body pose, PoseTrack surrounds the detected person with a bounding box and then declares the maxima of heatmaps as belonging together [2]. An alternative approach is to use heatmaps which provide richer supervision information compared to joint coordinates, by preserving spatial location information [220]. This information is ideal for training CNNs and has resulted in a growing interest in leveraging CNNs for the purpose of HPE [5,13,17,34,45,89,97,124,146,178,201,206,209]. Table 6 shows that the best performing body part detection-based method is achieved by Debnath et al. [34] with a PCKh@0.2 of 96.4%. As mentioned previously, this performance and the performance of other body part detection methods are higher than that of regression-based methods. Although body part detection methods have shown excellent performance, however, they are prone to estimating false positives [143].

#### *4.2.2 2D Multi-Person Pose Estimation*

The 2D MPPE aims to estimate the poses of multiple subjects from an input image and can be mainly categorised into the top-down, and bottom-up approaches. Furthermore, Table 6 shows that deep learning-based approaches are gaining in popularity due to better performance. Figure 3 presented a general taxonomy of 2D HPE whereas, [194] introduced a specific taxonomy of deep learning-based 2D MPPE. Deep learning-based MPPE can be further divided into one-stage or two-stage. The two-stage approaches can be further categorised into top-down and bottom-up methods similar to that of the general approach shown in Figure 3. Details of each of the sub-categories and approaches are provided in the following sub-sections.

##### *4.2.2.1 The 2D MPPE: Two-Stage Top-Down Approach*

The initial step of the two-stage top-down MPPE approach is person segmentation. This step aims to place each person inside an individual bounding box by using standard person detection methods such as Faster R-CNN [148], or Mask R-CNN [52]. Pose estimation can then be performed on each of the identified people where 2D coordinates for key points of each individual identified in an input image or video frame are estimated. The key points are then connected to create a 2D representation of the body pose of identified individual. Once the subjects are detected, the SPPE techniques can then be used to estimate the key points for each identified person and then to connect the key points to form an estimated pose [220]. The computing time in this approach is positively correlated with the number of individuals in image frame/s. Table 6 shows several recent studies that use this approach for 2D MPPE along with the performance outcomes.

An alternative approach to 2D MPPE is target representation that aims to estimate joint locations and can be further divided into two methods: coordinate-based and heatmap-based methods. Coordinate-based methods output two-dimensional coordinates of body key points. DeepPose was the first to use Deep Neural Networks (DNNs) for the problem of HPE [180]. DeepPose implements a cascade of DNN regressors and iteratively estimates joint coordinates via multi-stage refinements [194]. Coordinate-based methods such as DeepPose are intuitive, however, the method loses spatial information which can reduce the accuracy of the pose estimation. Heatmap-based methods, such as [124,169,179], manage to retain special information. Heatmap-based methods instead use the Gaussian heatmap as the learning target to encode the position of the joint [194]. Tompson et al. [179] presented the use of Gaussian heatmaps in HPE. Heatmap-based methods have been shown to reduce overfitting which is otherwise problematic for coordinate-based methods [17]. However, heatmap-based approaches are restricted by the resolution of the heatmap which is not consistent with the input image. Furthermore, obtaining key point coordinates is another limitation in the heatmap approach involving conversion of coarse coordinates in the heatmap to the input space, which influence the quantization error [194]. There are several methods available to mitigate this issue including: prior knowledge-based methods [124,215], soft argmax-based methods [98], and offset regression-based methods [58].

Another problem related to 2D MPPE is the variability in the scale of people in an image. To resolve the scale variance problem, multi-scale feature learning strategies, such as [180,201], have been developed. In DeepPose, the initial-stage regressor crudely estimates the position of key points in the full image while additional regressors take sub-images to learn the offset between the estimation and the true location [180]. Convolutional Pose Machines (CPM) [201] also follow a multi-stage process where heatmaps from previous stages are concatenated with low-level features to prevent vanishing gradients when training. The CPM reports a PCK of 90.5% [201] however, it indicates difficulties in separating the pose estimations when multiple subjects are in close proximity of each other.

Despite its success, the MPPE like other machine learning tasks can suffer from overfitting. To address this, Peng et al. [194] designed an adversarial data augmentation network based on Generative Adversarial Networks (GANs) [47], and reinforcement learning [111] and achieved improved performance by optimising them together. Toyoda [181] achieved improved HPE performance on images involving extreme and wild motions by implementing rotation augmentation. In addition to overfitting, person detectors often return redundant detections which may raise IoU scores in person detectors. This have been resolved by removing the redundant boundary boxes [39,141,216,217]. Papandreou et al. replaced IoU with OKS in Non-Maximum Suppression (NMS) to consider key points [141]. Zhang used soft-NMS which resets the redundant bounding boxes with low scores [216]. Parametric Pose NSM removes invalid key points generated by redundant detections in the post-processing [39].

*4.2.2.2 2D MPPE: Two-Stage Bottom-Up Approach*

In the bottom-up approach, key points are first identified in the image frame which is grouped into individual subjects and then joined together to form estimated poses [220]. Typically, the computing time for the bottom-up approach is lower than the top-down approach [220] as separate key points detection for each person is not necessary. As shown inTable 6, bottom-up methods indicate better performance [123]. An example of the bottom-up approach is DeepCut [144], which performs graph-based joint parsing in the form of an Integer Linear Program. Insafutdinov et al. adapt the deep Residual Network (ResNet) [124] for sliding window-based human body part detection (DeeperCut) [110] and DeepCut [109]. In contrast to graph-based approach, Cao et al. [20] perform greedy parsing on a tree structure to reduce complexity.

While the bottom-up approaches indicated satisfactory performance (up to 77.5% mAp in Table 6), they must overcome the foreground-background imbalance. This is an issue in which an infinite number of negative examples can be sampled, in comparison to a relatively small finite number of regions of interest, leading to an imbalance of negatives and positives [134]. OpenPose and PifPaf address the foreground-background class imbalance by using a greedy parsing algorithm to group key points for each individual [20,78]. Likewise, Simple Pose [84] is an alternative solution, where focal loss aims to tackle hard-to-learn pixels by balancing gradients of easy and hard samples.

*4.2.2.3 The 2D MPPE: One-Stage Approach*

In the one-stage MPPE approach, joint position estimation and the grouping of joints are performed in a single step. Recent work presented by Wang et al. [194] suggests that this approach harnesses the benefits of the two-stage approach whilst overcoming some of its limitations. The bounding box approach assigns key points to a subject only when the key points are located within the bounding box [52]. Associative embedding identifies the joint assignment as a tag regression task [124,169]. An alternative to the bounding-box method is the Offset-based method which aims to encode the connection between joints for key point assignments by aggregating offsets [126,140].

*4.2.3 The 2D Pose Estimation Summary*

Deep learning techniques have accelerated the performance in 2D HPE. Table 6 summarises several deep learning solutions indicating robustness for 2D SPPE including DeepPose [180], and the Stacked Hourglass Network [124]. The table also reports the examples of efficient approaches for 2D MPPE including RMPE [39] and OpenPose [20]. While addressing the task of 2D HPE, several studies have explored the use of temporal and special information to boost HPE [5,65,97,169,202,206]. The findings from MoDeep highlight the usefulness of motion features alone to outperform some traditional methods of HPE [65]. Gong et al. [46] suggested that motion cues such as optical flow provide useful information to aid in the extraction of key points, while the motion of rigid parts can aid in the identification of body joints. DeepFlow [202] utilises optical flow to better connect predictions between frames for a more continuous detection. Current methods involving optical flow calculate optical flow as a pre-processing step, however, machine learning solutions to calculating optical flow in real-time still need to be resolved. Motion-based features require at least two frames to compare, however, this is not feasible for most of the datasets available and therefore, further research in this area might be inhibited by the lack of relevant datasets.

This section presented both the body part detection and regression-based approaches to 2D SPPE. As mentioned earlier, the best performing body part detection-based method is achieved by Debnath et al. [34] with a PCKh@0.2 of 96.4%, and the best performing 2D SPPE in the regression category with the same metric reported was Unipose [5] with a PCKh@0.2 of 92.5%. Both approaches have their strengths and limitations as reported in previous subsections. This section has also reported on both the top-down and bottom-up approaches to 2D MPPE, Table 6 shows that the top-down approach has achieved the best performance with RSN [19] reporting mAP of 79.2%, whereas the best performance of a bottom-up method was OpenPose [20] with an mAP of 75.6%. However, the bottom-up approach has an advantage in speed as the input image goes through the network only once, whereas in the top-down approach each identified individual must go through the network, resulting in a slower performance [31].

Despite the success reported, there are additional limitations in the existing datasets including the larger datasets such as COCO [91] and Human3.6M [62] which provide a plethora of usual poses such as standing and walking. However, considerably fewer records of unusual poses such as upside-down and yoga poses are available in existing datasets. Therefore, generalise training of the models to detect and accurately estimate human poses may be a challenge [220]. Also, several works such as [114] report results based on a single dataset, often the COCO dataset without considering the data diversity. Reporting results from multiple datasets allows evaluating the model generalisation for alternative datasets.

### 4.3 3D Human Pose Estimation

The 3D HPE aims to extract an estimated 3D posture from an image and is a challenging compared to 2D HPE due to depth estimation of key points. It can be noticed from Table 4 that very limited datasets are available for 3D HPE as compared to 2D HPE datasets. Furthermore, these datasets are mostly limited to constrained, simulated and controlled environments such as motion capture laboratories. Very fewer datasets are captured in real-world environments with dynamic conditions such as varying backgrounds and moving objects, etc. For 3D HPE, there are two main types of input that include single view (i.e., images from only a single camera) or multi-view (i.e., images from multiple cameras placed at various angles). Like 2D HPE methods, 3D HPE can be categorised into SPPE and MPPE approaches as shown in the right-side tree of Figure 3 which is almost like that of 2D HPE is except for few differences. For instance, 3D HPE has top-level categories that include single view or multi-view. Likewise, 3D SPPE is classified as model-free or model-based which is not the case with 2D HPE. The 3D HPE methods will be described in further detail in the following subsection.

Table 7. A Comparison of 3D HPE Methods in Relation to Datasets used, Models and Performance Outcomes

| Name | Year | Dataset | SPPE/ MPPE | Approach | Model | Metric | Performance Score |
|---|---|---|---|---|---|---|---|
| [199] | 2021 | CSI Image | SPPE | Model-Based | DNN | P-MPJPE | 29.7mm |
| [107] | 2020 | MarCOnI, 3DPW MuCo-3DHP, MPI-INF-3DHP | MPPE | Bottom-Up | ResNet | MPJPE<br>3DPCK<br>AUC | 98.4mm<br>82.8%<br>45.3% |
| [150] | 2020 | Human3.6M, MPII LSPE, MuPoTS-3D | MPPE | Bottom-Up | CNN<br>RPN | 3DPCK | 74% |
| [193] | 2020 | MuPoTS-3D, CMU Panoptic | MPPE | Top-Down | RPN<br>ResNet | 3DPCK<br>MPJPE | 43.8%<br>30.5mm |
| [14] | 2020 | JTA, MuPoTS-3D, CMU Panoptic | MPPE | Top-Down | FPN | 3DPCK | 83.2% |
| [115] | 2019 | COCO, Human3.6M, MuPoTS-3D | MPPE | Top-Down | ResNet | 3DPCK | 31.5% |

| Name | Year | Dataset | SPPE/ MPPE | Approach | Model | Metric | Performance Score |
|---|---|---|---|---|---|---|---|
| [26] | 2019 | Human3.6M, HumanEva | MPPE | Top-Down<br>Model-Based | DNN<br>CNN | MPJPE<br>P-MPJPE | 42.9mm<br>32.8mm |
| [223] | 2019 | Human3.6M, HumanEva, MPI-INF-3DHP | SPPE | Model-Free | CNN<br>ResNet | MPJPE | 39.9mm |
| [83] | 2019 | Human3.6M, MPII, MPI-INF-3DHP | SPPE | Model-Free | CNN | MPJPE | 52.7mm |
| [4] | 2019 | Human3.6M, HumanEva, 3DPW | SPPE | Model-Based | ResNet | MPJPE | 77.8mm |
| [188] | 2018 | SURREAL, Unite the People | SPPE | Model-Based | CNN | MPJPE | 49mm |
| [135] | 2018 | UP-3D | SPPE | Model-Based | CNN | MPJPE | 59.9mm |
| [213] | 2018 | Human3.6M, CMU Panoptic | MPPE | Top-Down | DNN | MPJPE | 48mm |
| [87] | 2018 | MPII, Human3.6M | SPPE | Regression | CNN | MPJPE | 48.3mm |
| [108] | 2018 | MuCo-3DHP, Human3.6M, MuPoTs-3D | MPPE | Bottom-Up | CNN | 3DPCK | 75.2% |
| [106] | 2018 | Human3.6m, MPI-INF-3DHP | SPPE | Model-Based | CNN | MPJPE<br>3DPCK<br>AUC | 54.6mm<br>57.3%<br>28% |
| [100] | 2018 | 3DPW, Total Capture | MPPE | Top-Down, Model-Based | CNN | MPJPE<br>MPJAE | 26mm<br>12.1% |
| [149] | 2018 | Human3.6M | SPPE | Model-Based | CNN | MPJPE<br>N-MPJPE<br>P-MPJPE | 131.7mm<br>122.6mm<br>98.2mm |
| [99] | 2018 | MPII, Human3.6M, Penn Action, NTU | SPPE | Regression | CNN | MPJPE | 53.2mm |
| [151] | 2017 | Human3.6M, MPII, LSPE | MPPE | Bottom-Up | CNN<br>RPN | MJPE | 72.7mm |
| [224] | 2017 | MPII, MPI-INF-3DHP | SPPE | Model-Based | CNN | PCK<br>AUC | 69.2%<br>36.9% |
| [174] | 2017 | MPII, Human3.6M, HumanEva, KTH, LSP | SPPE | Model-Free | CNN | MPJPE | 69.7mm |
| [24] | 2017 | Human3.6M, MPII | SPPE | Model-Free | CPM | MPJPE<br>PCKh@0.5 | 82.7mm<br>88.5% |
| [117] | 2017 | HumanEva, Human3.6M | SPPE | Model-Free | CPM | MPJPE | 87.3mm |
| [142] | 2017 | HumanEva, Human3.6M, KTH | SPPE | Model-Free | CNN | MPJPE | 71.9mm |
| [177] | 2017 | Human3.6M | SPPE | Model-Based | CPM | MPJPE | 88.4mm |

Table 7 provides an overview of the existing 3D HPE methods including the relevant type and approach as per the taxonomy found in Figure 3. The bar chart in Figures S4 and S5 of the supplementary materials display the accuracy levels of the works presented in Table 7. Accuracies are reported in MPJPE (mm) and 3D PCK (%). Therefore, for MPJPE the smaller the bar, the higher the accuracy whereas, for 3D PCK the larger the bar the higher the accuracy**.** Figures S4 and S5 show that the best performing works are [100,115] which are described in the following subsections.

#### *4.3.1 The 3D Single Person Pose Estimation*

This subsection presents a range of 3D SPPE techniques that are mainly categorised as model-based or model-free as explained in the following subsections. Like 2D SPPE, 3D SPPE also performs person detection in the first step following image segmentation to isolate the identified person boundary within the image.

#### 4.3.1.1 Model-Based 3D SPPE

The model-based approaches for 3D SPPE, sometimes referred to as generative model approaches, which utilise a parametric body model template to estimate the human pose [25,155]. These approaches received significant attention in recent years [4,106,135,149,177,199,224]. By taking into account the human body's appearance, structure, and motions related to activities, model-based approaches can reduce the search space [143]. These approaches can further be classified into the kinematic (or skeleton) model and the volumetric model. The kinematic model, as employed in [24,100,108,142], include a set of joint positions and limb orientations to represent the human body structure [31]. It is the most widely used model for 3D HPE. The major advantage of the kinematic model is its simple representation using graph data structures [100]. The volumetric model, on the other hand, is used in 3D human reconstruction, and unlike the kinematic model, it can represent texture and shape information. BodyNet is one of the few methods as shown in Table 7 which makes use of the volumetric model [188] while deploying the Skinned Multi-Person Linear model [95]. Alternatively, the planar model [220] is also available for the 2D HPE estimation only. Another example of the model-based approach is presented by Zhou et al. [222] which validates regressed poses by employing a kinematic model to enforce orientation and rotation constraints by adding a kinematic layer in the framework to map the motion parameters to the joints. Nie et al. [125], and Lee etal. [82], use skeleton-LSTM to learn the depth information from global human skeleton features, leveraging joint relations and connectivity [31]. Chen and Ramanan [24] estimated the 3D postures from 2D kinematic models based on anthropometric, kinematic, and dynamic constraints.

#### 4.3.1.2 Model-Free 3D SPPE

In contrast to the model-based approach, the model-free 3D SPPE approach automatically learns the mapping between appearance and body pose leading to faster and precise estimations for certain actions [143]. While extensive literature is available for the model-free HPE [24,83,117,142,174,223], this approach has several limitations such as use of conventional methods (e.g., background subtraction) producing poor generalisation of poses [143]. In contrast, various existing studies take a direct estimation approach by inferring the 3D pose from 2D images without performing 2D HPE [86,87,142]. For instance, Sun et al. [87] proposed a structure-aware regression approach employing a bone-based representation reporting more stability than joint-based models. This work defines a compositional loss based on the long-range interactions between the bones [220]. Similarly, to convert the non-linear 3D coordinate regression problem into a more manageable discrete form, Pavlakos et al. [142] propose a volumetric representation.

The alternative to direct estimation is 2D to 3D lifting, where 3D HPE is inferred from a 2D pose estimation which is performed as an intermediate stage [220] and outperformed the direct estimation. Chen and Ramanan [24] used DNNs to lift 2D pose estimations to 3D pose estimations using nearest neighbour matching to implement a memorisation mechanism. Moreno-Neuger [117] formulates the 3D HPE task as a 2D to 3D distance matrix regression problem. This approach was found to better in handling the missing observations and allowed for the positions of non-observed key points to be hypothesised [117]. On the other hand, Zhou et al. [223] found that using the 2D to 3D lifting approach caused uncertainty due to the inherent ambiguities of the approach. They instead introduced an intermediate state Part-Centric Heatmap Triplets, to shorten the gap between the 2D estimation and the 3D interpolation [223]. The work was shown to generalise well to "in-the-wild" images where only weakly annotated 3D information is available [223].

### 4.3.2 The 3D Multiple Person Pose Estimation

This subsection presents a range of 3D MPPE techniques which can be categorised into top-down, or bottom-up approaches. Figure 3 indicates the position of 3D MPPE in the 3D HPE hierarchy.

#### *4.3.2.1 The 3D Top-Down MPPE Approaches*

Similar to the 2D top-down approach, 3D top-down MPPE also uses person detection in the first step to isolate and segment each person in an image. Then the root (i.e., the centre) of the identified individual, usually the pelvis, is localised which is then used for the root-relative 3D pose estimation, aligning all poses to the world coordinate [115]. Table 7 presents a variety of recently introduced 3D top-down models. For instance, Moon et al. [115] proposes a machine learning-based, camera distance-aware approach to 3D MPPE for single images. The work introduces RootNet, to estimate the camera-centred coordinates of each detected person, and PoseNet, to then estimate the root-relative 3D pose of each individual [115]. An alternative approach is LCR-Net that generates potential poses for each individual and refines these using a regressor [151]. However, LCR-Net lacks the generalisation to real-world images, and therefore LCR-Net++ was trained with synthetically augmented training data to address the issue [150]. A further alternative is proposed by Li et al. [193] which addresses the lack of a global perspective in the top-down approach by introducing a form of supervision, Hierarchical Multi-person Ordinal Relations (HMOR). HMOR captures the body part and joint level semantic while maintaining global consistency by encoding interaction information as ordinal relations of depth and angles hierarchically.

#### *4.3.2.2 The 3D Bottom-Up MPPE Approach*

The first step in the 3D bottom-up approach is the estimation of joint locations and depth maps. The estimated joint locations and body parts are then associated with the individual subjects, according to the root depth and part relative depth [220]. One of the major issues with the bottom-up approach is the occlusion occurrence due to hidden-view of joint locations that has been addressed in several works. As an example, Mehta [109] presents Occlusion-Robust Pose-Maps (ORPM) which produces full-body pose estimations even under strong partial occlusion. ORPM uses body part association to allow for the inference of 3D poses without explicit bounding boxes. Similarly, Zhen et al. [219] regresses a set of 2.5D representations of body parts and then from this, reconstructs the 3D poses using a depth-aware part association algorithm by reasoning about inter-person occlusion and bone-length constraints. Mehta [107] also presented the SelecSLS Net architecture which infers 2D and 3D pose encodings for visible joints which then reconstructs 3D body poses based on the results of a person detector. Finally, refinement is performed for temporal stability and kinematic skeleton fitting.

### *4.3.3. The 3D Pose Estimation Summary*

To summarise, the 3D HPE is more complex problem as compared to 2D HPE. The reliance on motion capture data implies that most datasets are captured in a laboratory setting. It would be useful to acquire the dataset from RGB and motion capture data acquisition in varying real-world environments while making use of the portable motion capture hardware. Without a dataset captured in a real-world environment, it will remain a challenge to develop a generalised model for the real-world scenarios (i.e., images and environments). Figure 5 summarises the distribution of HPE datasets utilisation grouped by HPE category. Datasets which account for less than 2% of the distribution are categorised as ‘other’. The MPII, COCO, and LSPE datasets are mostly cited and referenced in 33.33%, 21%, and 19.3% of the 2D HPE literature reviewed respectively. Although the FLIC dataset has been referenced in 11.4% of all the 2D HPE literature reviewed, it has only been cited once in the last five years. This is likely because authors have shown a preference for creating custom-made datasets, rather than collating existing images from movies and other similar online resources. For 3D HPE, the literature contains a wider variety of datasets as compared to 2D HPE. The most cited dataset of the 3D HPE literature reviewed is Human3.6m and is referenced in 34.6% of the articles reviewed.

Table 7 shows that the best performing 3D SPPE method is Wi-Mose [199], a model-based approach, reporting an MPJPE of 29.7mm. This is better than the best performing model-free method, HEMlets Pose [223] with MPJPE of 39.9mm. However, model-free approaches indicated faster performance than model-based approaches [143]. Likewise, model-free approaches are sensitive to appearance changes while model-based approaches struggle to extract reliable body models from gait sequences [210] the subject of the next section. Table 7 also indicate HMOR [193] being outperforming 3D MPPE method (a top-down method), with an MPJPE of 30.5mm. This was better than the best performing bottom-up method, LCR-NET [151], which reported an MPJPE of 72.7mm. As with 2D MPPE, the top-down approach provides the best accuracy. However, bottom-up approaches deliver faster performance because they run the image through the model or network only once as compared to top-down approaches executing for each identified individual.

Researchers have only recently begun to harness the deep learning for 3D HPE [101,129] however, it is likely to follow 2D HPE in terms of its popularity and acceleration of accuracy improvement. As discussed in Section 4.2.3, recent works have initiated the exploration of the use of temporal features such as optical flow for 2D HPE. To the best of the authors' knowledge, this has not yet been explored for 3D HPE, however, this may present opportunities to improve the estimation. The success of top-down 3D HPE is highly dependent on the recent developments in person identification and SPPE. As with the 2D HPE, the computational complexity may increase significantly with increasing numbers of individuals in an image. Likewise, due to the cropping performed by the person detectors, a significant amount of background information may eliminated by the top-down approaches leading to inaccurate depth estimations. In contrast, bottom-up approaches benefit from linear increments in computational time with respect to the number of individuals identified in an image.

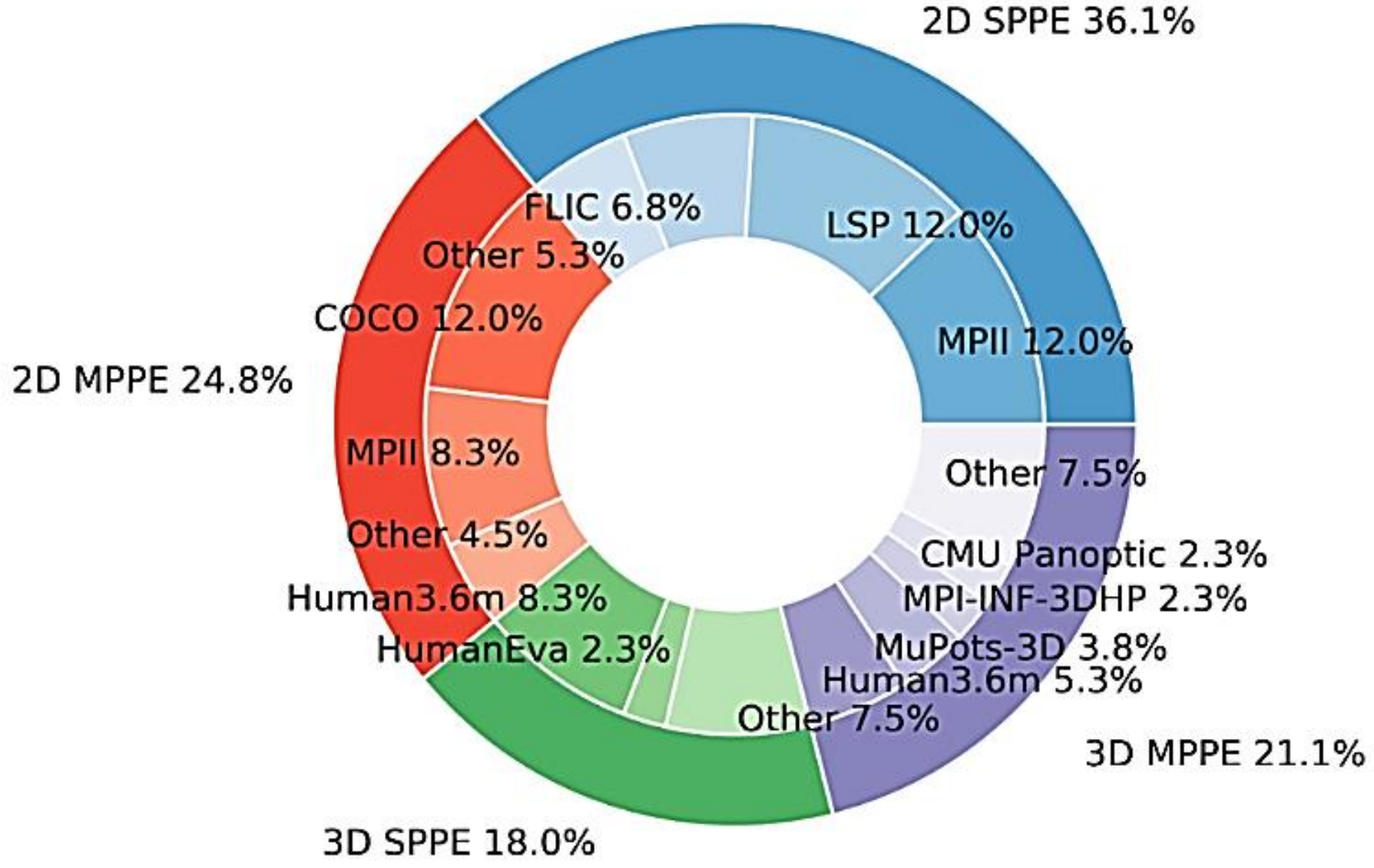

Figure 5. Distribution Summary of the Datasets Usage Grouped by the Related HPE Category

## 5 GAIT IDENTIFICATION

Gait refers to a person's manner of walking and has shown to be a unique biometric identifier for the person identification. Concerning person identification, gait recognition has significant advantages over other identification methods such as facial recognition and fingerprint identification. For instance, it is possible to identify subjects via their gait at a distance or a low resolution when other methods (e.g., face matching) may fail under these circumstances [139,173]. Studies have shown that it is difficult for a person to disguise their unique gait as this usually impedes movement [130]. Gait analysis is also of great importance to physiotherapists and for rehabilitation where it can be used as a tool to identify and aid in the treatment of unhealthy gait [185]. Physiotherapy is costly and labour intensive and therefore, by using automatic gait analysis, individuals at home can accelerate their recovery by completing exercises at home with automated feedback.

Person identification or re-identification via gait analysis has the advantages of being unobtrusive. It does not require the cooperation of the subject, can be measured at a distance, is unique to each individual, and cannot be easily faked or concealed [120]. However, person identification via gait does have its limitations such as variation in gait due to illness, age, and emotional states [35,160]. Gait can also vary based on the walking surface, types of clothing, objects being carried and clutter in the environment [120]. Accurate HPE and a skeleton-based approach may provide a method of analysing the gait in a way that is not affected by clothing, environmental clutter or personal appearance.

Gait analysis can be performed using a variety of sensors including cameras, pressure sensors and inertial sensors. The use of pressure sensors and inertial sensors is less well documented for the application of person identification as compared to the computer vision approach. This is likely because the use of such sensors requires some contact and consent from the subject. The use of computer vision approaches has the advantage of being more passive and can perform the gait analyses of a subject from a distance, without their knowledge, though consent would be required unless performed by concerned authority (e.g., police and security agencies). A selection of gait datasets created with the intention of supporting person identification is shown in Table 8 along with Table 9 presenting the related studies utilising these datasets.

Table 8. A Comparison of Existing Gait Datasets for Person Identification

| Dataset Name | No. Subjects | Description | Dataset Name | No. Subjects | Description |
|---|---|---|---|---|---|
| CASIA-datasetA [196] | 20 | Outdoor | KinectREID [138] | 71 | Indoor |
| CASIA-datasetB [196] | 124 | Indoor | I-LIDS [182] | 119 | Indoor |
| SOTON (large) [163] | 114 | Indoor | MARS [221] | 1,261 | Outdoor |
| TUM-GAID [55] | 305 | Indoor | PKU [176] | 18 | Outdoor |
| OU-MVLP [172] | 10,307 | Indoor | PRID2011 [53] | 983 | Outdoor |

The datasets presented in Table 8 have several limitations. A common limitation is a relatively low number of participants such as in [138,176,196]. Likewise, many datasets are limited to indoor environments only, such as in [55,138,163,172,182,196]. Although Mars [221] contains a variety of participants in real-world environments, the recorded videos are inconsistent in terms of the viewing angle and distance from the participants and does not label the angles or distances which affects the estimated key point locations required for the skeleton-based approaches. On the other hand, some datasets such as [182] are no longer available for the public use.

Table 9. Performance Analysis of Gait-based Person Identification

| Ref | Approach | Dataset | No. Subjects | Rank-1 CMC rate / mean CCR (Unless otherwise stated) |
|---|---|---|---|---|
| [198] | Discriminative Selection in Video Ranking. | PRID2011, iLIDS-VID,<br>HDA+ | 200<br>300<br>83 | 40%<br>39.5%<br>54.3% (5fps), 52% (2fps) |
| [121] | Multi-modal feature fusion of 3D soft biometric cues. | Vislab KS20 | 20 | 39.6% |
| [122] | Context-aware ensemble fusion. | Vislab KS20 | 20 | 74.67% (No Context)<br>82.33% (Cross-context) |
| [15] | Marker-less feature extraction. | iLIDS | 20 | 92.5% |
| [8] | Fisher Linear Discriminant + Maximum Margin Criterion (MML). | CMU Graphics MOCAP | 464 | 75%-85% (ROC) |

The literature provides numerous approaches to person identification using gait, Table 9 summarises several of the cutting-edge solutions [120]. Many studies used the traditional silhouette-based approach to gait analysis for person identification [190,196]. While other works take a skeleton-based approach to gait recognition for person identification using RGB images [173,210] or depth images [121]. For instance, Yao et al. [210] used Skeleton Gait Energy Images (SGEIs), and the results showed improvements over traditional Gait Energy Images (GEIs) when participants used additional clothing items such as long coats. Alternatively, several works use a CNN approach to extract the gait features [173,207]. As an example, Xie et al. [207] make use of additional information with cross-modal embedding layers to help focus on each person's prominent distinctions. Similarly, Upadhyay et al. [184] perform HPE and analyse gait to identify people with the use of deep learning. There are alternative to computer vision techniques for identifying people via their gait such as use of the footstep sounds [22,162]. Likewise, studies have identified pressure sensors on the floor as being useful to identify people [136,145].

To analyse gait for person identification, rehabilitation, or other purpose, it is necessary to extract distinguishing features, which will provide relevant clues regarding a person's gait. These features may include walking velocity, cadence, stride duration, stride length, and stride width [153]. A traditional method of analysing gait involved computing a gait cycle into a single Gait Energy Image (GEI) [96,173]. GEI preserves dynamic and static information of a gait cycle such as the appearance and shape of the body and the variation and frequency of the gait phase [96]. The two major approaches for gait feature extraction from RGB images are the traditional silhouette approach [155] and the more recent skeleton approach [173]. The silhouette approach involves extracting the silhouette or outline of the human body from the scene, while the skeleton approach involves extracting an HPE skeleton from the image. Study presented in [173] shows a comparison of the different gait representation methods of a person in the CASIA-B [212] gait dataset at different timesteps [173]. Each row depicts the same frames as the original RGB image, silhouette image, and 2D skeleton pose, respectively [173]. It is observed that the RGB image includes background details, whereas the silhouette image extracts the individual from the background. Furthermore, study reported that the silhouette image is affected by the accessories worn by the individual whereas, the skeleton pose contains only estimated joint locations and representations of their connections.

Figures S6 and S7 of the supplementary materials present a comparison of the gait identification accuracies for both the silhouette-based and the skeleton-based approaches. Both approaches include examples of works that achieved over 90% accuracy. Figures S6 and S7 show that the highest accuracy work belongs to the silhouette-based approach. Both approaches and the individual works are explored in sections 5.1 and 5.2. The remaining section mainly focuses on gait feature extraction techniques from RGB images as well as alternative methods, which have been reported in the literature.

### 5.1 Silhouette Approach

Silhouette extraction is an important step in many computer vision tasks such as gait recognition [176], body part segmentation [10], and HPE [193]. Object detection methods such as the Haar Cascade Classifier [191] can be used first to detect human body within the image frame. The identified body is then isolated from the background image, leaving only the binary silhouette of the body as shown in [173].

Once a silhouette image is achieved, various feature extraction methods can be used to analyse gait, including GEI, Gait Energy Image on Depth Data (depth-GEI), Gait Energy Volume (GEV), and Depth Gradient Histogram Energy Image (DGHEI) [55]. GEI assumes that all gait information can be captured in a single gait cycle and therefore averages the data captured over one full cycle [50]. This method discards a lot of information from each cycle; however, it may remove some of the noise that is captured. From RGB data alone, silhouettes can be of low quality and may not accurately capture the outline of the body [55]. An alternative to GEI is Depth-GEI, which can overcome some of these limitations by using depth data to produce a higher quality GEI. A further alternative to GEI is GEV, which is an extension to GEI to average the three-dimensional binary voxel volumes [167]. Another feature extraction method, DGHEI, uses the concept of averaging feature vectors of each frame of a gait cycle, as with the standard GEI. DGHEI also uses the edges and depth gradients available in the depth data [54]. Luo et al. [96] propose Accumulated Frame Difference Energy Image (AFDEI) as a method of considering time within a GEI. AFDEI is calculated by combining forward frame difference image with backward frame difference image. The accumulated frame difference energy image is obtained using the weighted average method.

There are a few studies such as CASIA-B [212] providing pre-processed silhouettes as part of their dataset to aid other works. Alternative studies, [172] provide only silhouettes instead of RGB images, to eliminate the need to share identifiable images of participants. As an example, Chauhan [23] provides such a dataset containing silhouettes only.

Table 10. Comparison of Silhouette-based Approaches to Gait Recognition

| Ref | Year | Features / Extraction Methods | Models / Classifiers | Accuracy |
|---|---|---|---|---|
| [203] | 2022 | GEI | GAN | 51.78% |
| [176] | 2019 | GEI | RankSVM | N/A |
| [190] | 2018 | GEI | K-Means Clustering | 92% |
| [96] | 2017 | GEI, AFDEI | Nearest Neighbour (NN) | 88.7% |
| [161] | 2016 | GEI | Convolutional Neural Network (CNN) | <= 89.7% |
| [204] | 2015 | GEI | CNN | <= 90% |
| [81] | 2009 | View Transformation Model (VTM) based on GEI | Linear Discriminant Analysis (LDA) | <= 90% |
| [50] | 2006 | GEI | Euclidean Distance Classifier (EDC) | <= 89% |
| [94] | 2005 | Body shape, gait stance shape | K-Means Clustering | 78% |
| [196] | 2003 | Principal Component Analysis (PCA) | NN, Nearest Neighbour with respect to class exemplars (ENN) | 82.5% |

Table 10 provides an overview of silhouette-based approaches to gait analysis and identification, including the features used for recognition and the accuracy of the gait recognition. As evident in Table 10, GEI is a popular approach for extracting gait features from silhouette images. Recent studies explore the deep learning techniques for gait recognition from GEI [161,204]. GEINet [161] achieved recognition accuracy of 89.7% when using GEIs in a CNN designed for gait recognition. The model was trained over the OU-ISIR dataset [64]. Although the dataset contains 4007 subjects with diverse characteristics, it also has limitations such as limited range of view variations. Likewise images were captured in a green environment to enable easier silhouette extraction [64,161]. Similarly, Wu et al. [204] uses CNN while utilising adjacent video frames as input from which 4,096-dimensional frame-level features are extracted. The features are then forwarded to a Multilayer Perceptron (MLP) for the gait recognition producing up to 90%accuracy [204]. However, this model ignores motion between frames that could be helpful to improve recognition rates.

Obtaining a silhouette image can also be used as a starting point for other methods. For example, Jalal et al. [66] identified key points in the silhouette to obtain a skeleton model with a body part detection rate of 90.01% [66]. Although this approach reported successful detection, it does not report the accuracy of the pose estimation. Furthermore, the model was not trained to overcome occlusion. Similarly, Barnard et al. [10] trained Hidden Markov Models with labelled synthetic images to label silhouette body parts which is further used to annotate data automatically. However, this model also does not produce pose estimation for the estimated body parts.

While silhouette-based gait analysis indicates satisfactory outcomes in several recent works, a common problem with the silhouette model is that it tends to differ when taken from different angles. To resolve this, View Transformation Model (VTM) transforms gait features, as in [81,176], from multiple gallery views to the probe view for allowing the evaluation of gait similarity. However, it is not possible to use the VTM approach when the viewing angle is unknown or is not included in the predefined views. Tian et al. [176] address this with their View-Adaptive Mapping (VAM) where the viewing angle of the gait sequence is estimated and a Joint Gait Manifold (JGM) is used to find the optimal manifold between the probe data and relevant gallery data to evaluate gait similarity [176]. Alternatively, Wen et al. [203] transform the GEI to 90° normal state using GANs. A related issue occurs when cameras are placed in an elevated position, such as on a hill or attached to a streetlight, the top-down view can often cause self-occlusion where one body part hides another. Verlekar et al. [190] propose a solution to this problem, by using shadows as a means of obtaining a less occluded silhouette. Though this partially solves the problem of occlusion, it has other associated issues. For example, the solution does not perform well without appropriate lighting conditions [190]. Also, slopes and environments can distort shadows and obstacles (and their shadows) can occlude shadows [190].

In summary, the literature reports success when using the silhouette approach as summarised in Table 10. However, many of the datasets used has various limitations that mainly include insufficient variety of participants, variety in real-world environments or varied viewing angles, and comprising the pre-processed silhouette images or GEI without the original RGB data. Likewise, the silhouette approach itself also has some limitations. For example, clothing and accessories can alter the appearance of the silhouette and image quality may also affect the silhouette produced.

### 5.2 Skeleton Approach

It can be noticed from [173] that the outline of the binary silhouette images can be easily affected by clothing and accessories. A possible solution is the skeleton approach that aims to produce a consistent skeleton pose estimation regardless of what the participant is wearing or carrying. The motion of human skeletons provides a significant amount of information about the subject, their activities, and gait. To implement the skeleton-based approach, a skeleton for each individual must be extracted using HPE, as previously discussed, with visualisation examples shown in [173].

Table 11. A Comparison of Existing Skeleton-based Approaches to Gait Recognition

| Ref | Year | Features / Extraction Methods | Models / Classifiers | Accuracy |
|---|---|---|---|---|
| [147] | 2021 | CAGEs | Neural Network (NN) | <=92% |
| [173] | 2021 | Graph | GCN | <=87.7% |
| [160] | 2021 | Graph | AT-GCN | 92.2% |
| [88] | 2020 | Joint angles, limb lengths, and joint motions | CNN | <=76.1% |
| [211] | 2019 | SGEI | CNN | <=82.33% |
| [139] | 2019 | Pose + 3D Face | Decision Tree | 92.3% |
| [210] | 2018 | SGEI | CNN | <=92.09% |
| [11] | 2018 | Pose | TGLSTM | 98.4% |
| [121] | 2017 | Pose | NN | 96.67% |
| [1] | 2014 | Joint Angles | K-Nearest Neighbour (KNN), MLP | 80% |
| [9] | 2012 | 3D Pose | K-Means Clustering | 43.6% |

Table 11 presents a range of skeleton-based approaches to gait including the feature extraction methods, models and classifiers used, in addition to the accuracy achieved. One skeleton-based feature extraction approach is SGEI, a GEI that is generated from HPE skeletons rather than silhouette images which have been discussed earlier. It was first proposed by Yao et al. [210], achieving a recognition accuracy of 92.09%, as shown in Table 11, and has recently grown in popularity. SGEI starts with real-time HPE which identifies 13 key points in the limbs and the torso. The key points are then used to calculate or extract features such as thigh swing and step period. SGEI has shown to be more robust to changes of clothes when compared to other methods such as GEI [210,211].

An alternative skeleton-based feature was recently proposed by Rao et al. [147] introducing Constrastive Attention-based Gait Encodings (CAGEs) as a distinct feature representing the gait effectively. It performs 3D HPE firstly, to gain a 3D skeleton from which gait features can be extracted. This approach differs from other existing work, such as [139], as it provides a self-supervised method to encode discriminative gait features from unlabelled 3D skeletons [147]. Unlike [11,51,88], this approach does not require hand-crafted features or prior data annotations for supervised gait representation learning [147]. This method achieved gait recognition accuracy of 92% [147] as shown in Table 11. Despite the advantages and accuracy of this model, it is limited in several ways. For example, the dataset used to train the model is relatively small [147]. Similarly, the model considers only skeletons generated using high-quality data such as depth data and therefore, it is unknown whether this model would generalise well in uncontrolled environments such as the outdoors [147]. Finally, the estimated skeletons are modelled as body-joint sequences on different dimensions which may not capture underlying relations between the joints [147].

Earlier skeleton-based works, such as [76,92], relied on traditional CNNs which are not optimised for the graph-structured data, though recognition accuracy of 92.2% was achieved [160]. More recently, several studies have used graph-structured data when working with HPE data for tasks such as activity recognition and gait recognition. As an example, Graph Convolutional Networks (GCNs) has been proposed as a generalisation of CNNs to work with graph-structured data [16]. The GCNs are capable of automatically learning both spatial and temporal patterns from data. Early use of GCNs focused on human activity recognition [164,208]. Studies have shown that GCNs can be used for the problem of gait recognition [160,173]. For example, GaitGraph [173] reported 87.7%, 74.8%, and 66.3% recognition accuracy on the CASIA-B [212] dataset with normal walking, with a bag, and with a coat or jacket respectively. The reported accuracy is lower than the state-of-the-art silhouette approaches, such as the 92% reported in [190]. However, the silhouette-based approaches use appearance to aid recognition which may not be considered as true gait recognition [173]. The work presented in GaitGraph [173] suggests that true temporal gait features can be used to recognise gait alone.

An alternative graph-based learning approach, Time-based Graph Long Short-Term Memory (TGLSTM), is proposed in [11]. It is inspired by Recurrent Neural Networks (RNN) and consists of Long Short-Term Memory (LSTM) nodes alternating to full connected layers. TGLSTM achieved the highest gait recognition accuracy (98.4%) as reported in Table 11. In contrast to RNNs, the TGLSTM model works in a frame-by-frame manner while learning the skeleton joint features as well as the information extracted from the changes in the adjacency matrix over time [11]. With deep learning, the model can exploit temporal data, including how new connections are formed and old ones broke to learn long short-term dependencies. However, in this method each skeleton is a polygonal approximation which is not a natural-looking skeleton.

Another alternative graph-based learning approach known as Attention Enhanced Temporal Graph Convolutional Network (AT-GCN) is presented by Sheng and Li [160]. The model produced a recognition accuracy of 88.9% on the TUM-GAID [55] dataset and 92.2% on their dataset, EMOGAIT [160], which is not currently available to the public. This solution attempts to identify participants via their gait, predict future frames of the gait, and finally predict emotions, such as sadness, from the gait data [160]. Another study presented by Pala et al. [139] suggests that combining the skeleton-based approach with facial recognition techniques can outperform the individual methods. In this work, the authors combined the traditional 2D skeleton approach with a 3D model of the face using a depth sensor and reported an accuracy of 92.3%. However, this model indicated appropriateness for only certain distances between the camera and the participant.

In summary, the accuracy of the skeleton-based approaches shown in Table 11 is almost like that of the silhouette-based approaches shown in Table 10. Table S1 of the supplementary materials provides a statistical comparison of both approaches where the t-test outcome indicates no significant difference between the means of both approaches. This is despite the fact that silhouette images also contain visual clues which can help identify individuals [173], however, such visual clues may not be available outside of controlled environments, for example, varying lighting conditions or clothing worn. Therefore, a comparison of the approaches using a more dynamic real-world dataset would be more informative. Some models such as [173] indicate good performance by utilising gait features only, however, the authors in [139] suggest that combining gait and appearance may provide higher recognition accuracy. A detailed discussion on the alternative approaches, technologies, and applications of gait analysis is provided in Section 1.6 of the supplementary materials.

## 6 DOMAINS AND APPLICATIONS

HPE and real-time gait analysis have been employed in various domains and have a substantial impact via diverse applications within these domains. Gait has been used as soft biometric data, which may prove useful for security and governance bodies such as law enforcement for criminal identification or missing person identification. Accurate HPE and gait analysis could also be used in real-time analysis within sports for commentary or coaching. Another potential domain where this technology could have an impact in healthcare, particularly in detecting abnormal gait for disease diagnosis and rehabilitation. This subsection will introduce several such domains and applications.

### 6.1 Person Identification

There are many scenarios where accurate person identification is required including border control, access control, criminal identification, authentication systems, safety systems such as construction and environmental, missing person identification, and many more. Biometric identifiers are distinctive physiological characteristics of a person which may be used to label and describe or identify a person. Examples of biometric identifiers include fingerprints, DNA, iris recognition and facial recognition etc. As discussed in earlier sections, gait has been considered as an important biometric identifier. The use of DNA, fingerprints and iris recognition are difficult to perform without the knowledge or cooperation of the individual that requires identification. The literature shows that humans and machines can recognise people by their gait [63,120]. The use of facial recognition and gait analysis can be performed covertly, at a distance and without the knowledge of the individual. As discussed previously, facial recognition has several limitations, specifically in the case of low-quality images such as blurry images, background noise, or when the face is hidden etc. On the other hand, gait analysis has an advantage over facial recognition specifically because it is extremely difficult for someone to disguise gait without inhibiting movement. Also, the need for high-resolution images is not a major requirement here.

### 6.2 Healthcare

Abnormal gait can be identified and corrected through physiotherapy. Physiotherapy is costly, labour intensive, and involves long waiting times for treatment. Gait rehabilitation could be aided and accelerated with regular home exercise, guided by an automated feedback solution [75]. An example is proposed by Ropars et al. [152], which used motion capture data to analyse the range of motion in the shoulders of participants. The method can assess shoulder hyperlaxity, a main risk factor for shoulder instability. Similarly, [12] used motion capture technology to diagnose knee injuries. Likewise, it may be possible to create an alternative diagnostics system using HPE in place of the motion capture suite. In addition to diagnosis and gait rehabilitation, [77] shows that gait analysis provides the potential identification of gait imbalance and fall prediction, such solutions may enable quick response to incidents, for example, fallers in care homes.

### 6.3 Sports and Sports Coaching

Sport is an integral part of many people's lives and plays a significant role in individuals' quality of life. It is important to play sports with the correct posture and techniques to perform to the highest standards, avoid injuries and avoid future health issues [73]. The analysis of gait has been proven as an effective method of quantifiably comparing athlete's performance, particularly for sports such as sprinting [36]. It presents the opportunity to allow clinicians and sports coaches to analyse the biomechanics of athletes and to aid them in becoming more effective in their sport. Similarly, for sports such as boxing, judo, snooker, javelin, cricket, and football, whole body posture is important for optimal performance. HPE can offer the opportunity to automatically analyse athletes' performance to aid with sports coaching.

One example of automated coaching is presented by Tharatipyakul et al. [175] that uses HPE to compare the tai chi technique between a user's video and a trainers video. They were able to use this comparison to verify the accuracy of the user's movements for their tai chi training. To estimate the user's performance, the average angular difference of the shoulder, hip, upper and lower arms, as well as upper and lower legs between the trainer and the participant was calculated. The authors did not implement any automated coaching in this solution. Similarly, Takeichi et al. [171] used HPE to analyse the performance of runners, using six metrics; speed, step frequency, step length, vertical oscillation, trunk angle, arm swing angle and leg swing angle. Again, the authors did not implement any automated feedback for the participants. Similarly, Wang et al. [195] used HPE to track athletes while they ski. They implement a novel HPE model by refining coarse heatmaps generated by a basic image HPE model. This approach successfully estimates unusual body poses such as an upside-down person, which is especially important for sports such as skiing and snowboarding. In the current implementation, feedback is relatively simple where poses are classified as either good or bad and users are alerted when they perform a bad pose. However, prescriptive feedback on how to improve the bad poses is not currently provided.

The reviewed studies indicate that a variety of work already published in sports and athlete analysis while there is a significant opportunity for researchers to build on this work, especially for sports coaching. If it is possible to track an athlete's body pose, then it must be possible to provide qualitative feedback on their pose in a real or near-real time manner. For example, if a system can analyse a boxing guard based on the angle and position of their arms, it should be a trivial problem to provide feedback on whether the boxer should raise or lower their guard, bring their hands closer together or further apart. During the literature review, the authors are not aware of any solutions which provide this type of feedback.

## 7 DISCUSSION

Many datasets for both HPE and gait identification have been reviewed as part of this survey. Two of the most popular and most practical datasets are COCO[91], and Human3.6m [62]. Both datasets contain a large number of instances with high quality labelling, however, they are several limitations in both datasets. For example, COCO lacks a variety of unusual poses and comprises a single viewpoint, whereas Human3.6m is restricted to a laboratory environment and contains limited diversity in subjects. These limitations are typical of the existing datasets currently available. Similarly for the gait datasets, TUM-GAID [55] and SOTON HiD [183] provide a large number of subjects along with multiple gait cycles per subject. However, like other datasets, they are limited in terms of the diversity of environments, viewpoints, and subject clothing. Detailed recommendations for an improved dataset are provided as future directions in section 8.

In relation to existing HPE models and approaches, Table 6 indicated that he top-down approach has achieved the best 2D HPE performance with RSN [19] reporting mAP of 79.2%. However, it has associated limitations, for example, the computational cost is directly affected by the number of subjects in an image. Similarly, this method often suffers from redundant person identifications. Furthermore, the top-down approach to 3D HPE is limited by the same challenges as that of the 2D problem. The top-down approach often implements SPPE as the second step after identifying person in an image. The SPPE regression methods currently perform lower as compared to the body part detection methods. For example, the best accuracy was reported by Debnath et al. [34] with a PCKh@0.2 of 96.4%. Although body part detection methods have shown excellent performance, however, they are prone to estimating false positives [143].

The 3D HPE is a more challenging problem than 2D HPE due to estimation for the depth dimension in addition to the 2D coordinates of key points. Wi-Mose [199], a model-based approach was the best performing 3D SPPE method, reporting an MPJPE of 29.7mm. This is a better performance than the best performing model-free method, HEMlets Pose [223] with MPJPE of 39.9mm. However, model-free approaches have indicated faster performance than model-based approaches [143]. Likewise, model-free approaches are sensitive to appearance changes while model-based approaches struggle to extract reliable body models from the gait sequences.

The gait identification (as mentioned previously in section 5) can be categorised into the silhouette-based approaches and the skeleton-based approaches. Most significantly, the skeleton-based approaches have been indicating the gait analysis independent of personal appearance, unlike the traditional approaches which do not guarantee the identification using purely gait data [139,210]. However, the statistical analysis provided in Table S1 of the supplementary materials does not show significant difference ($p=0.608$) between the mean accuracy of both approaches at present. The authors are not aware of any current attempts towards implementing the gait identification using 3D HPE. Such a solution might potentially offer the additional information and features, such as stride width, which is not possible using 2D HPE.

## 8 CONCLUSIONS AND FUTURE WORKS

This study has surveyed a plethora of approaches and applications of gait analysis using computer vision techniques such as HPE. This survey has also provided a comprehensive review of the datasets and technology available to support gait analysis and person identification. This includes a comprehensive review of HPE approaches and solutions which are imperative to the implementation of gait analysis techniques following a skeleton-based approaches. A comparison of both the traditional silhouette-based approaches and the more recent skeleton-based approaches has been provided and the advantages of the skeleton-based approaches have been highlighted. This survey should provide the reader with sufficient direction to aid them in exploring and developing skeleton-based gait analysis and identification which will prove useful in a variety of domains and applications. It has also identified several limitations in the literature which provide ample opportunity for researchers to make significant contributions in a vast range of domains. Potential ideas for future work are as follow:

- The 3D HPE-based gait analysis for person identification that would allow for additional gait features to be extracted, such as stride width. . Currently, there has been limited progress in this area.
- HPE-based crime or violence detection and classification. By training machine learning models to identify criminal actions, it would be possible to automatically alert law enforcement agencies about a crime in real-time. There are sporting datasets available that may aid in the training of models to identify actions such as punching, kicking, and throwing which would be key indicators of violence in the real world. The authors are unaware of any specific crime-based datasets available.
- Optic flow has been shown to provide useful data for the gait analysis and person identification. Current methods require optic flow to be pre-calculated during pre-processing stages. It would be beneficial if machine learning models could be trained to automatically detect and calculate the optic flow in real or near-real time. The authors are not aware of any attempt to implement this solution. Similarly, research is necessary to further explore the use of optic flow as a method of analysing gait.
- Generally, the current methods of HPE are dependent on machine learning models learning to identify the key points. This has been shown to cause some limitations such as the difficulty dealing with key point occlusion. Developing a solution without this dependency would provide a useful alternative approach. This can be achieved using body part segmentation and localisation however, may require the development and/or labelling of a dataset.

There are ample opportunities to train machine learning models to effectively coach individuals for a range of sports and martial arts based on data derived from HPE and gait analysis.